%% file: main.tex
\documentclass[11pt]{article}

\usepackage[preprint]{acl}

\usepackage{times}
\usepackage{latexsym}

\usepackage[T1]{fontenc}

\usepackage[utf8]{inputenc}

\usepackage{microtype}

\usepackage{inconsolata}

\usepackage{graphicx}
\usepackage{booktabs}
\usepackage{amssymb}
\usepackage{amsmath}
\usepackage{CJKutf8}
\usepackage{makecell}
\usepackage{multirow}
\usepackage{xcolor}
\usepackage{caption}
\usepackage{needspace}
\usepackage[most]{tcolorbox}
\usepackage{dblfloatfix}

\newtcolorbox{samplebox}[1]{
  width=\linewidth,
  enhanced,
  colback=black!3,
  colframe=black!78,
  coltitle=white,
  colbacktitle=black!78,
  fonttitle=\bfseries\small,
  title={#1},
  boxrule=0.5pt,
  arc=1.6mm,
  left=4pt, right=4pt, top=4pt, bottom=4pt,
  before skip=5pt, after skip=2pt
}

\newcommand{\ct}[1]{\texttt{#1}}

\title{\textbf{CSWBench}: Evaluating Context-Sensitive Code-Switching in Creative Writing}
\author{
    \normalfont\normalsize
    \begin{tabular}{@{}c@{}}
    {\large
    \textbf{Zewen Jiang}\textsuperscript{1,2}\enspace
    \textbf{Yahan Li}\textsuperscript{3}\enspace
    \textbf{Chenlu Guo}\textsuperscript{4}\enspace
    \textbf{Diana Inkpen}\textsuperscript{5}\enspace
    \textbf{Yuan Wu}\textsuperscript{1}\enspace
    \textbf{Yi Chang}\textsuperscript{1,6,7}}
    \\[0.3em]
    \textsuperscript{1}School of Artificial Intelligence, Jilin University, China \\
    \textsuperscript{2}College of Software, Jilin University, China \\
    \textsuperscript{3}Xingchen AGI Lab, China Telecom Artificial Intelligence Technology (Beijing) Co., Ltd., China \\
    \textsuperscript{4}Bilibili Inc., China \quad
    \textsuperscript{5}University of Ottawa, Canada \\
    \textsuperscript{6}Engineering Research Center of Knowledge-Driven Human-Machine Intelligence, MOE, China \\
    \textsuperscript{7}International Center of Future Science, Jilin University, China
    \\[0.3em]
    jiangzw1324@mails.jlu.edu.cn
    \quad
    liyh181@chinatelecom.cn
    \quad
    guochenlu01@bilibili.com
    \\
    Diana.Inkpen@uottawa.ca
    \quad
    \{yuanwu, yichang\}@jlu.edu.cn
    \end{tabular}
}
\hypersetup{
    pdftitle={CSWBench: Evaluating Context-Sensitive Code-Switching in Creative Writing},
    pdfauthor={Zewen Jiang, Yahan Li, Chenlu Guo, Diana Inkpen, Yuan Wu, Yi Chang}
}

\begin{document}
\maketitle
\begin{abstract}
Large language models (LLMs) have demonstrated multilingual capabilities across diverse language generation and understanding tasks. Yet effective code-switching requires more than multilingual proficiency: models must decide when, where, and why to switch languages in ways that fit the surrounding context. This context-sensitive ability remains insufficiently evaluated. We introduce \textbf{CSWBench}, a benchmark for evaluating context-sensitive code-switching in creative writing through two complementary settings: Focused Narrative and Extended Narrative. We further propose \textbf{CSWEval}, a theory-informed framework with two complementary levels: \textbf{event-level evaluation} of Licensing, Contextual Appropriateness, and Functional Gain, and \textbf{whole-text evaluation} of task-specific quality---overall literary and narrative quality in Focused Narrative and quality emphasizing multilingual organization in Extended Narrative. Evaluating twelve LLMs, we find that models exhibit meaningful but uneven code-switching competence across narrative contexts. Model differences are particularly pronounced in multilingual literary integration, while multilingual-focused dimensions show stronger difficulty effects overall than the general narrative dimensions; narratively salient switches are associated with stronger contextual and functional realization; and event-level CS quality is moderately associated with whole-text performance. The artifact is available at \url{https://github.com/TPSJASPER/CSWBench}.
\end{abstract}

\begin{table*}[t]
\centering
\setlength{\tabcolsep}{4.3pt}
\renewcommand{\arraystretch}{1.1}
\small

\resizebox{0.95\textwidth}{!}{%
\begin{tabular}{lcccccc}
\toprule
\textbf{Benchmark} & \textbf{Task} & \textbf{Granularity} & \textbf{Generation} & \textbf{CS Choice} & \textbf{CS Orch.} & \textbf{Salience} \\
\midrule
LinCE~\cite{aguilar-etal-2020-lince} & CS NLP & Token / Utterance & $\times$ & $\times$ & $\times$ & $\times$ \\
CodeMixBench~\cite{yang-chai-2025-codemixbench} & CS Understanding & Token / Text & $\triangle$ & $\times$ & $\times$ & $\times$ \\
ChiEngMixBench~\cite{yang2026chiengmixbench} & CS Generation & Sentence Pair / Text & $\checkmark$ & $\checkmark$ & $\times$ & $\times$ \\
LitBench~\cite{fein-etal-2026-litbench} & Creative Writing & Story Pair & $\checkmark$ & $\times$ & $\times$ & $\times$ \\
WritingBench~\cite{wu2025writingbench} & Writing & Whole Text & $\checkmark$ & $\times$ & $\times$ & $\times$ \\
\textbf{CSWBench} & \textbf{Creative Writing} & \textbf{Event + Whole Text} & $\checkmark$ & $\checkmark$ & $\checkmark$ & $\checkmark$ \\
\bottomrule
\end{tabular}
}
\caption{Comparison of \textbf{CSWBench} with representative code-switching and writing benchmarks. Task denotes the evaluation setting; Granularity lists representative assessment units. Generation denotes evaluation of generated text; CS Choice indicates whether switching decisions are evaluated against contextual conditions; CS Orch. indicates whether multilingual behavior is assessed across a text; and Salience indicates whether the narrative importance of individual switches is considered. $\checkmark$ denotes explicit coverage, $\triangle$ partial coverage, and $\times$ an aspect not evaluated. NLP = natural language processing; CS = code-switching; Orch. = orchestration.}
\label{tab:benchmark-comparison}
\end{table*}

\section{Introduction}

Code-switching (CS) refers to the alternation between two or more languages or language varieties within the same interaction or discourse~\cite{gardner-chloros-2009-code,bullock-toribio-2009-cambridge}. Its meaning depends on social and conversational context, serving identity expression, relationship negotiation, interlocutor accommodation, and affective communication~\cite{gumperz-1982-discourse,defina-2007-code,ladegaard2018codeswitching,giles-coupland-coupland-1991-accommodation}. Such context-sensitive language choices are also common in contemporary digital communication, where they serve a range of communicative and pragmatic functions~\cite{bali-etal-2014-borrowing,rudra-etal-2016-understanding,rijhwani-etal-2017-estimating}.

\begin{figure}[t]
\centering
\includegraphics[width=\columnwidth]{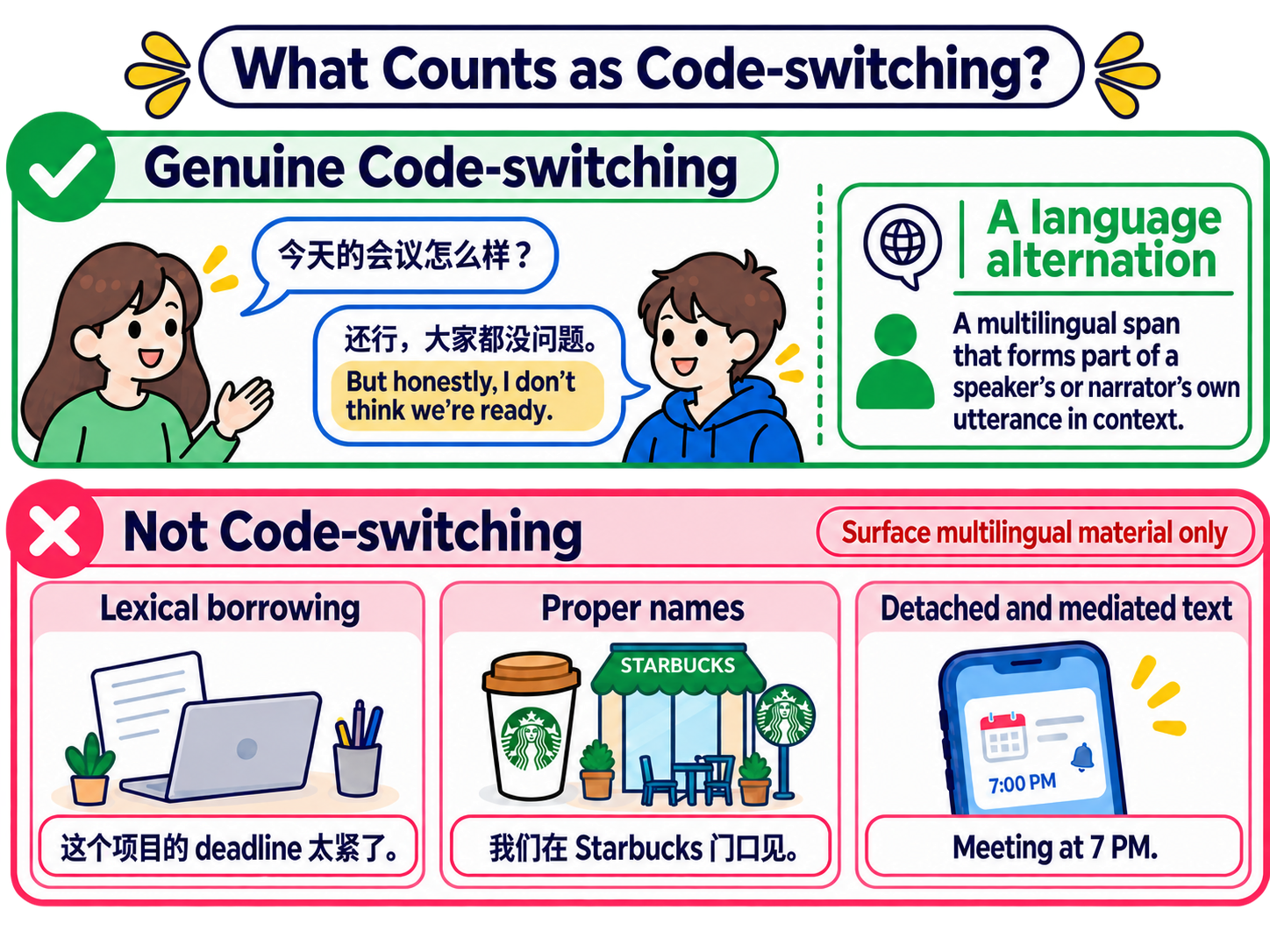}
\caption{Illustrative distinction between genuine code-switching and surface multilingual material. The examples distinguish meaningful language switches from English expressions that do not constitute code-switching (CS). A genuine switch involves English use attributable to current speaker or discourse voice as part of an ongoing interaction or narrative. By contrast, proper names, conventionalized borrowings, and quoted or displayed English without active uptake are excluded. The distinction depends on how English participates in discourse, rather than on the mere presence of English words.}
\label{fig:what-counts-cs}
\end{figure}

Existing code-switching evaluations have largely examined how models process code-switched inputs through downstream NLP tasks, including language identification, named entity recognition, question answering, and natural language inference~\cite{aguilar-etal-2020-lince,khanuja-etal-2020-gluecos}. LLM-era studies have extended this paradigm to translation, reasoning, and truthfulness~\cite{huzaifah-etal-2024-evaluating,yang-chai-2025-codemixbench,mohamed-etal-2026-lost}. Meanwhile, generated CS has been evaluated for fluency, naturalness, and semantic fidelity~\cite{heredia-etal-2025-conditioning,aksut-etal-2026-german}, or against naturally occurring switching patterns~\cite{sterner-teufel-2025-minimal,winata-etal-2026-understand}. Yet whether LLMs can make contextually appropriate, functionally motivated code-switching decisions remains comparatively underexplored.

Creative writing foregrounds the context-sensitive nature of code-switching, as literary language choices can be deliberately staged within a narrative~\cite{grutman-2024-stage}. Characters' linguistic resources, relationships, and circumstances provide grounds for judging individual switches. For example, a character who speaks English cautiously might repeat an interlocutor's English phrase to challenge a claim. Whether this switch is convincing depends on the character's established language abilities, its fit with the ongoing exchange, and the contribution it makes to the disagreement. Across a narrative, such choices can also shape characterization, relationships, and narrative development~\cite{gardner-chloros-weston-2015-codeswitching,baldo-deganutti-2024-narrative}. Creative writing thus makes the contextual grounding, local functions, and broader narrative roles of code-switching observable and evaluable.

To fill this gap, we introduce \textbf{CSWBench}, a benchmark for evaluating whether LLMs can use code-switching as a context-sensitive language choice in creative writing. Drawing on spontaneous and controlled Chinese--English source material, \textbf{CSWBench} transforms screened bilingual records into structured narrative prompts that preserve their interactional grounding while adding limited narrative scaffolding through model-assisted construction, programmatic checks, and recorded screening and refinement. The benchmark comprises two complementary settings: \textbf{Focused Narrative}, which foregrounds individual code-switching choices within their local narrative contexts, and \textbf{Extended Narrative}, which places greater emphasis on the organization and development of multilingual behavior across the narrative as a whole.

Drawing on sociolinguistic and literary theories of code-switching, we develop \textbf{CSWEval}, a multi-level evaluation framework that assesses individual switching events in terms of Licensing, Contextual Appropriateness, and Functional Gain, together with task-specific whole-text quality, including multilingual organization in Extended Narrative. Evaluating twelve contemporary LLMs reveals three recurring patterns. First, multilingual literary integration clearly differentiates models, while multilingual-focused dimensions show stronger difficulty effects overall than the general narrative dimensions. Second, switches that carry greater narrative importance also tend to be more successfully realized in their immediate context and function. Third, local CS quality is moderately associated with whole-text performance.

Our contributions are threefold:

\noindent(i) We introduce \textbf{CSWBench}, a benchmark for evaluating context-sensitive code-switching in creative writing through complementary Focused and Extended Narrative settings that foreground local multilingual choices and whole-text multilingual organization, respectively.\\
\noindent(ii) We develop \textbf{CSWEval}, a theory-informed, evidence-grounded evaluation framework that distinguishes Licensing, Contextual Appropriateness, and Functional Gain at the event level from task-specific quality at the whole-text level.\\
\noindent(iii) We conduct a systematic evaluation of twelve contemporary LLMs, providing empirical evidence on where models succeed and fail in using code-switching as a context-sensitive narrative resource.

\section{Related Work}

\noindent\textbf{Computational Evaluation of Code-Switching.} Computational studies have evaluated code-switching competence through several paradigms. Input-oriented work examines downstream prediction and language understanding, as in LinCE~\cite{aguilar-etal-2020-lince} and GLUECoS~\cite{khanuja-etal-2020-gluecos}, as well as LLM evaluations of translation and reasoning~\cite{huzaifah-etal-2024-evaluating,yang-chai-2025-codemixbench} and comprehension~\cite{mohamed-etal-2026-lost}. Generation-oriented work evaluates controllability and fluency~\cite{mondal-etal-2022-cocoa,yong-etal-2023-prompting}, semantic fidelity and human judgments~\cite{aksut-etal-2026-german}, and correspondence with naturally occurring switching patterns~\cite{winata-etal-2026-understand}.

Recent work has increasingly examined the conditions underlying individual language choices. Minimal-pair evaluations compare model preferences with those of bilingual speakers~\cite{sterner-teufel-2025-minimal}, while OLA examines pragmatic cues~\cite{oh-etal-2026-ola} and ChiEngMixBench considers community-specific switching conventions~\cite{yang2026chiengmixbench}. A broader survey situates these developments within multilingual evaluation across modalities~\cite{sheth-etal-2026-beyond}. Our work extends this focus to the contextual and narrative roles of language choices in creative writing.

\noindent\textbf{Creative Writing and Narrative Evaluation.} Creative-writing evaluation treats narrative quality as a multidimensional construct, assessing coherence and reader engagement~\cite{chhun-etal-2022-human}, creativity and literary quality~\cite{chakrabarty-etal-2024-art}, and task-specific writing quality and human preferences~\cite{wu2025writingbench,fein-etal-2026-litbench}. Long-form evaluation further examines character consistency and event retention~\cite{wang-etal-2025-generating}, as well as cross-scene continuity~\cite{wu-etal-2025-shifting}. Building on these approaches, our work examines the local functions of code-switching and the development of multilingual behavior across characters, relationships, and narrative states.

\noindent\textbf{Sociolinguistic and Literary Perspectives on Code-Switching.} Code-switching is theorized through complementary perspectives on interactional, social, and narrative functions. Interactional accounts emphasize discourse position~\cite{auer-1995-pragmatics} and speaker alignment~\cite{gumperz-1982-discourse,goffman-1979-footing}. Social theories relate language choice to interlocutors, social distance~\cite{bell-1984-audience,giles-coupland-coupland-1991-accommodation}, identity, and stance~\cite{bucholtz-hall-2005-identity}. Literary perspectives examine code-switching as a staged narrative resource~\cite{grutman-2024-stage} shaping characterization, relationships, and textual organization~\cite{gardner-chloros-weston-2015-codeswitching,baldo-deganutti-2024-narrative}.

\begin{figure*}[t]
\centering
\includegraphics[width=\textwidth]{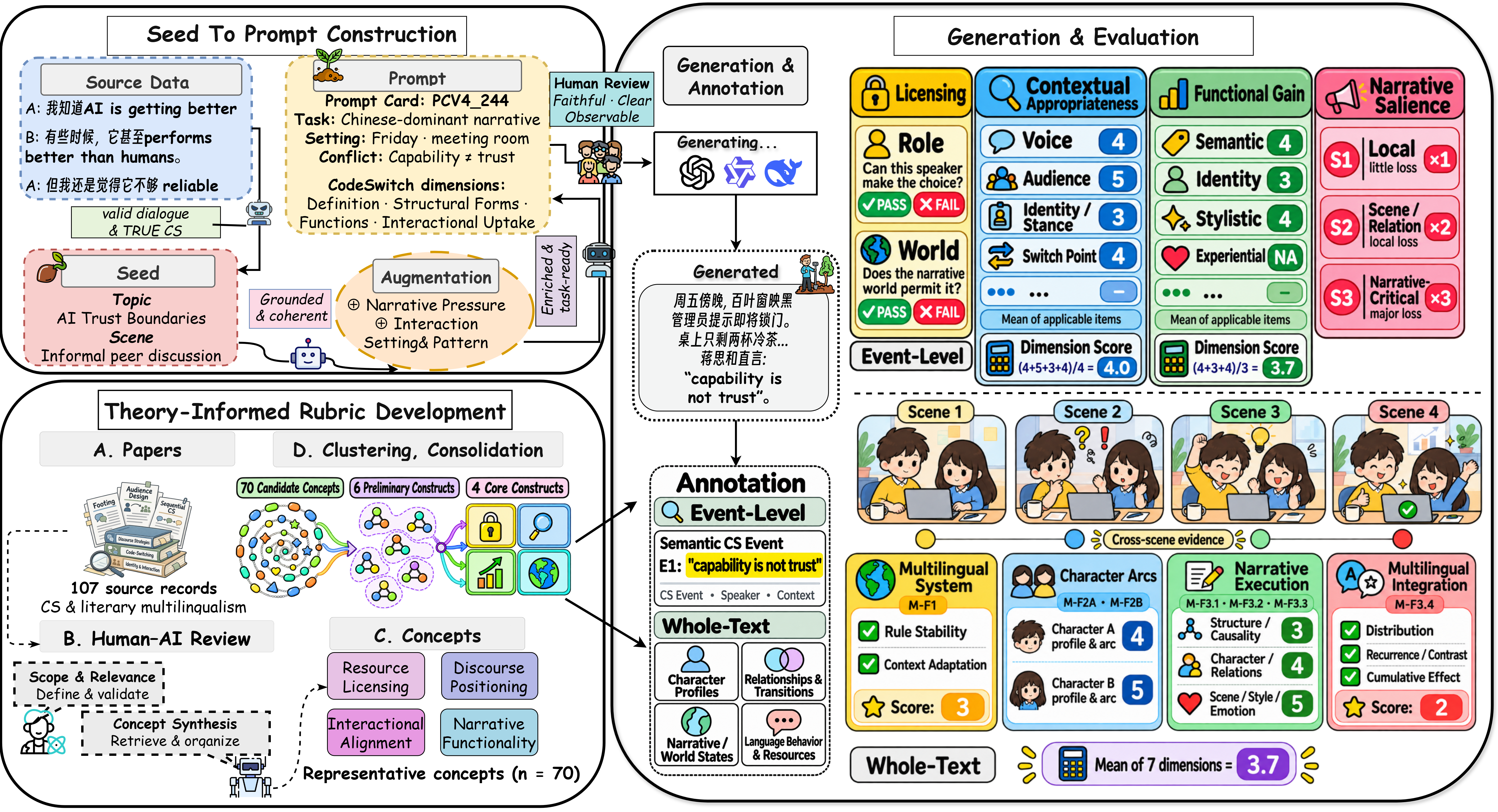}
\caption{Overview of the \textbf{CSWBench} construction and \textbf{CSWEval} evaluation pipeline. Source records are screened and transformed into Seed-conditioned prompts; CSWEval defines event-level and whole-text assessment; and twelve LLMs generate Focused and Extended narratives for evaluation. Event-level dimensions include Licensing (D1), Contextual Appropriateness (D2), and Functional Gain (D3). The illustrated whole-text dimensions are for Extended Narrative; text fragments and scores are schematic.}
\label{fig:pipeline}
\end{figure*}

\section{Method}

Starting from bilingual interactions in ASCEND~\cite{lovenia-etal-2022-ascend} and SwitchLingua~\cite{xie2025switchlingua}, we build structured \textbf{Seeds} and transform them into open-ended prompts for narrative generation. As illustrated in Figure~\ref{fig:pipeline}, our pipeline comprises three stages: data and prompt construction, development of \textbf{CSWEval}, and generation and evaluation.

\subsection{Data Source and Seed Construction}
\label{sec:data-seed}

We construct \textbf{CSWBench} from ASCEND's naturally occurring Chinese--English conversations~\cite{lovenia-etal-2022-ascend} and SwitchLingua's controlled CS corpus~\cite{xie2025switchlingua}. After source-specific normalization, we sample 3,500 records and screen for interpretable interactional context and attributable CS. We exclude records containing only lexicalized borrowings, proper names, detached multilingual material, or quoted or mediated material without active uptake by the current discourse voice. Screening and sensitive-content filtering retain 3,009 records for Seed construction. Normalization and coverage statistics appear in Appendix~\ref{app:data}.

For each retained source sample, GPT-5.6-Sol~\cite{openai-2026-gpt56} constructs a structured \textbf{Seed} by extracting source-grounded information about participants, topics, situations, and language-use cues. Limited narrative scaffolding, including plausible participant relations, interaction patterns, and narrative pressure, provides a usable creative-writing context without prescribing individual switches. Constructions whose additions are inconsistent with the source context are discarded. DeepSeek-V4-Pro~\cite{deepseek-2026-v4} then validates source fidelity and scaffolding consistency, yielding 596 validated Seeds.

The final sets maximize contextual diversity across the source pool, jointly covering all 14 recorded scene domains and all 11 interaction patterns. Extended retains every scene domain; Focused omits one. Screening, construction, and selection details appear in Appendix~\ref{app:construction}.

\subsection{Theoretical Foundations of \textbf{CSWEval}}
\label{sec:theory-foundations}

Our constructs draw on interactional approaches to CS and literary multilingualism. Interactional accounts treat language choice as situated social action shaped by participants, local conditions, and unfolding discourse~\cite{gumperz-1982-discourse,auer-1995-pragmatics}: ``Why this language, by this speaker, here and now?'' Literary multilingualism treats multilingual fiction as a deliberately shaped resource rather than a transparent reproduction of bilingual speech~\cite{delabastita-grutman-2005-fictional,grutman-2024-stage,baldo-deganutti-2024-narrative}. Together, these perspectives motivate our view of code-switching in creative writing as a situated multilingual choice organized as a narrative resource.

Building on these foundations, we conducted an integrative theory review using database searches and citation tracing. The review retained \textbf{107} scholarly works and identified 70 candidate evaluative concepts, which were consolidated into six preliminary constructs. Four inform the quality framework, while the remaining two guide evidence requirements and narrative-salience weighting. Search and construct-development procedures are provided in Appendix~\ref{app:theory}.

These constructs ground \textbf{CSWEval}: three concern individual switching events and one concerns the narrative as a whole.

\paragraph{Event-level constructs.} Three constructs underpin event-level evaluation. (i) Participant--world resource grounding asks: Can this participant make this choice? It concerns whether a multilingual choice is supported by the participant's established linguistic resources and represented narrative world, and is operationalized as \textbf{Licensing} (D1). (ii) Situated interactional coherence asks: Does it fit here? It concerns whether the choice is coherent for relevant voice, participants, interactional context, and narrative moment, and is operationalized as \textbf{Contextual Appropriateness} (D2). (iii) Functional multilingual contribution asks: What does this multilingual choice contribute here? It concerns the semantic, relational, identity-related, stylistic, or experiential contribution made by the multilingual choice within the event, and is operationalized as \textbf{Functional Gain} (D3). At the event level, these constructs assess whether a switch is licensed by the speaker and narrative world, appropriate to its local context, and functionally meaningful.

\paragraph{\mbox{Whole-text construct.}} For Extended Narrative, Whole-text Multilingual Orchestration asks whether multilingual choices form a coherent, developing pattern across characters, relationships, scenes, and narrative progression. Unlike the event-level constructs, \mbox{it is judged over the whole narrative}, considering how multilingual choices recur, contrast, and evolve across the text. Operationally, this construct is represented by seven scored whole-text dimensions in Extended Narrative, spanning multilingual organization and general narrative quality.

\subsection{Task and Prompt Construction}
\label{sec:task-prompt}

From the pool of 596 validated Seeds, we construct task-specific subsets for Focused Narrative and Extended Narrative using coverage-first sampling over source and contextual characteristics. The same Seed may support both settings.

\subsubsection{Seed-to-Prompt Construction}

For each selected Seed, GPT-5.6-Sol constructs Seed-conditioned content and renders candidate prompts in four-part templates covering a fixed creative-writing instruction, character and scene information, narrative-development guidance, and task-specific requirements. Focused separates character and scene fields; Extended combines them and adds narrative-development instructions. Representative Prompt Cards appear in Appendix~\ref{app:examples}.

The two settings differ in generation requirements. Focused and Extended Narrative target approximately 700 and 4,000 characters, respectively. Focused Narrative establishes a narrative situation, while Extended Narrative requires a developing narrative world with evolving characters, relationships, narrative states, and multilingual behavior. Both settings leave specific code-switching choices and story development open to the model, yielding 855 candidate prompts for subsequent validation.

\subsubsection{Prompt Validation}

Candidate prompts undergo deterministic checks for completeness, formatting, prohibited information, and redundancy, followed by two independent human reviews for Seed fidelity, task clarity and satisfiability, CS observability, and creative openness under setting-specific criteria. Disagreements receive third-rater adjudication; correctable problems undergo targeted revision and re-screening. The final set retains 807 of 855 prompts: 407 Focused and 400 Extended, derived from 546 distinct Seeds, with 261 shared across tasks.

The final sets retain diversity in scene domains, participant relations, interaction patterns, and preferred styles. The paired reviews yield Gwet's AC1 = .975 for Focused and .838 for Extended; screening details are reported in Appendix~\ref{app:validation}.

\subsection{\textbf{CSWEval}: Evaluation Framework}
\label{sec:csweval}

\textbf{CSWEval} evaluates generated narratives at two complementary levels: individual code-switching events and the narrative as a whole. The constructs developed in Section~\ref{sec:theory-foundations} are operationalized through task-specific, evidence-grounded rubrics: binary checks applied at the event level for D1, aggregated at the article level as $5\times$ the pass rate, and 1--5 ordinal rubrics with dimension-specific anchors for D2, D3, and whole-text quality. The complete dimensions, subdimensions, and task-specific scoring schemes are summarized in Appendix~\ref{app:rubric}.

\subsubsection{Code-Switching Event Annotation}

Before evaluation, every contiguous Latin-script span is treated as a candidate CS span. Candidates are classified using the operational CS definition and grouped into events, fixing their boundaries before quality scoring. For Extended Narrative, we prepare character and language profiles, relationship and narrative-world records, and Licensing evidence available at each event for event-level Licensing and longitudinal whole-text evaluation. Annotation rules appear in Appendix~\ref{app:annotation}. Human audits examine event discovery and evidence-record fidelity (Appendices~\ref{app:human} and~\ref{app:evidence-fidelity}).

\subsubsection{Task-Specific Multi-Level Evaluation}

Both settings evaluate event-level Licensing, Contextual Appropriateness, and Functional Gain but differ in scoring granularity and whole-text rubrics. Focused scores each applicable Contextual Appropriateness and Functional Gain subdimension per event and aggregates them into dimension scores. Extended uses these subdimensions as rubric guidance and assigns one overall score per dimension.

At the whole-text level, both settings score multiple dimensions separately and aggregate them into an overall whole-text score, but use different task-specific rubrics. Focused Narrative assesses overall literary and narrative quality. Extended Narrative evaluates seven whole-text dimensions: multilingual narrative-system stability (M-F1), the bilingual profiles and language development of two focal characters (M-F2A and M-F2B), three general narrative dimensions covering structure, character development, and prose quality (M-F3.1--M-F3.3), and multilingual literary integration (M-F3.4).

\subsubsection{Narrative Salience and Score Aggregation}

Not all identified code-switching events contribute equally to narrative development. Narratological accounts of eventfulness distinguish events by the extent and consequences of the changes they introduce into a developing narrative~\cite{schmid-2010-narratology,huhn-2014-event}. Each judge independently assigns a Narrative Salience level (S1, S2, or S3) to each previously identified code-switching event during evaluation. These levels reflect increasing narrative consequence and correspond to aggregation weights of 1, 2, and 3, respectively. Contextual Appropriateness and Functional Gain are then aggregated across identified code-switching events using these salience weights. Human-agreement results are effectively unchanged under alternative salience-weighting schemes: equal weighting and both milder and more concentrated alternatives yield correlations within approximately .01 of the reported 1:2:3 scheme.

Whole-text evaluation is computed independently from the event-level branch. Given $K$ task-specific whole-text dimensions with scores $W_k$, the article-level whole-text score is

\begin{equation}
W_{\mathrm{article}} =
\frac{1}{K}\sum_{k=1}^{K} W_k.
\end{equation}

When a dimension contains multiple scored subdimensions, the applicable subdimension scores are aggregated by averaging them before article-level reporting. Within each narrative setting, we then summarize model performance with the equal-weight composite $\Sigma=D1+D2+D3+\mathrm{Whole}$ (/20), while retaining the individual dimensions for more detailed analysis.

\section{Experiment Setup}

We evaluate twelve contemporary large language models spanning proprietary and open-weight model families. All twelve models were evaluated with shared decoding settings (temperature $0.7$, top-$p$ $0.7$, and a 30,000-token output limit); the complete model roster and generation-side reasoning configurations are reported in Appendix~\ref{app:provenance}.

The evaluation set contains 407 Focused Narrative prompts and 400 Extended Narrative prompts. The generation design assigns one narrative per model and prompt, under the configuration recorded in Appendix~\ref{app:setup}. We evaluate the generated narratives using \textbf{CSWEval} (Section~\ref{sec:csweval}), with generation-model identities hidden from the evaluators. Each narrative is independently evaluated by GPT-5.6-Sol and DeepSeek-V4-Flash~\cite{deepseek-2026-v4}. Licensing (D1) is judged event by event through the role- and world-level licensing checks, and the two judges' article-level scores are averaged for model-level reporting.

\input{appendix_support/main_overall_performance.tex}

\section{Results and Analysis}
\label{sec:results}

\phantomsection\label{sec:overall-performance}

Table~\ref{tab:overall-performance} summarizes model performance in both settings. GPT-5.5 leads Focused Narrative ($\Sigma=15.537$), whereas GPT-5.6-Sol leads Extended Narrative ($\Sigma=17.767$). Both remain top-ranked under evaluator-family exclusion and alternative aggregations (Appendices~\ref{app:self-preference} and~\ref{app:aggregation-sensitivity}).

\subsection{\textbf{CSWEval} Validity}
\label{sec:validity}

\subsubsection{Human Validation}
\label{sec:rater-agreement}

Three human raters evaluated 180 Focused and 144 Extended outputs (Kendall's $W=.718$ and $.810$, respectively). Comparing human-mean and dual-judge article scores yields ICC(A,1) = .864 and Spearman's $\rho=.863$ for Focused, versus ICC(A,1) = .816 and $\rho=.804$ for Extended.

Two independent bilingual annotators audited candidate classification (108 narratives) and event grouping (48 narratives). Human audits closely agree with automatic event discovery (mean post-stratified candidate-classification F1: .963/.960; event-grouping F1: .944/.998 for Focused/Extended; Appendix~\ref{app:human}). Changing the Prompt constructor and event annotator affects absolute scores, broadly preserving rankings (Appendix~\ref{app:pipeline-sensitivity}).

\begin{figure*}[t]
\centering
\begin{minipage}[t]{0.456\textwidth}
\centering
\includegraphics[width=\linewidth]{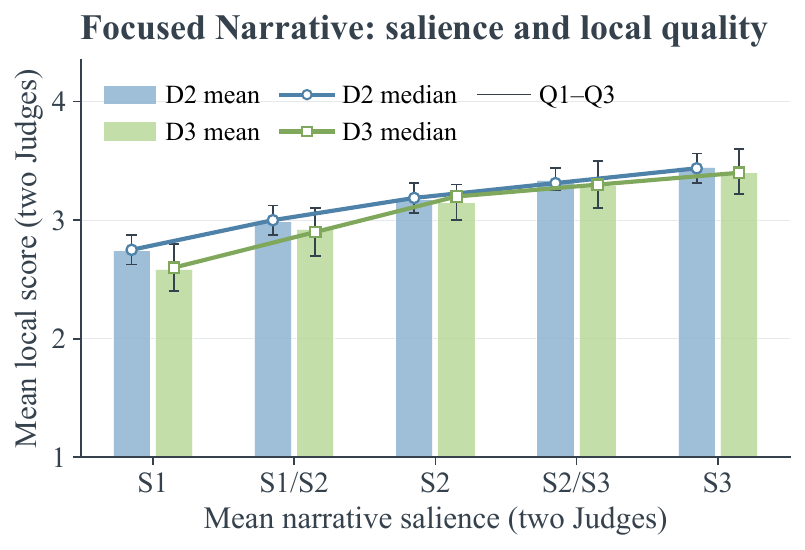}
\caption*{(a) Focused Narrative salience}
\end{minipage}
\hfill
\begin{minipage}[t]{0.456\textwidth}
\centering
\includegraphics[width=\linewidth]{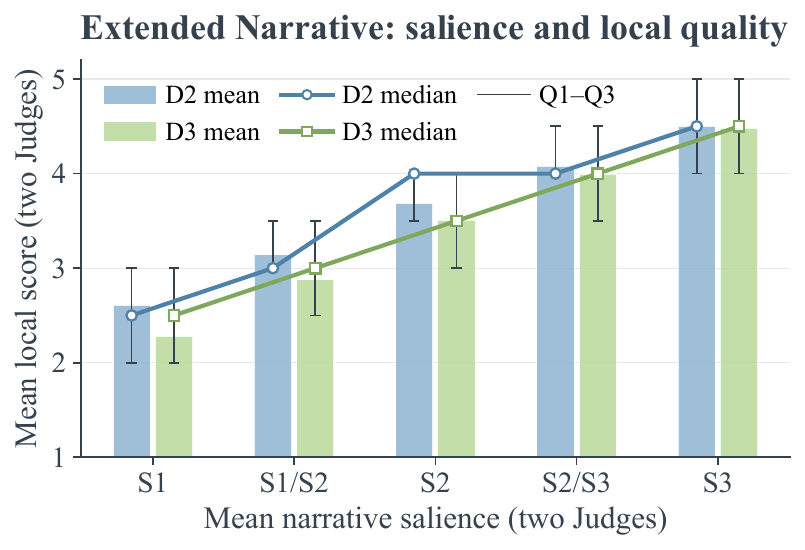}
\caption*{(b) Extended Narrative salience}
\end{minipage}

\vspace{5pt}
\includegraphics[width=0.95\textwidth]{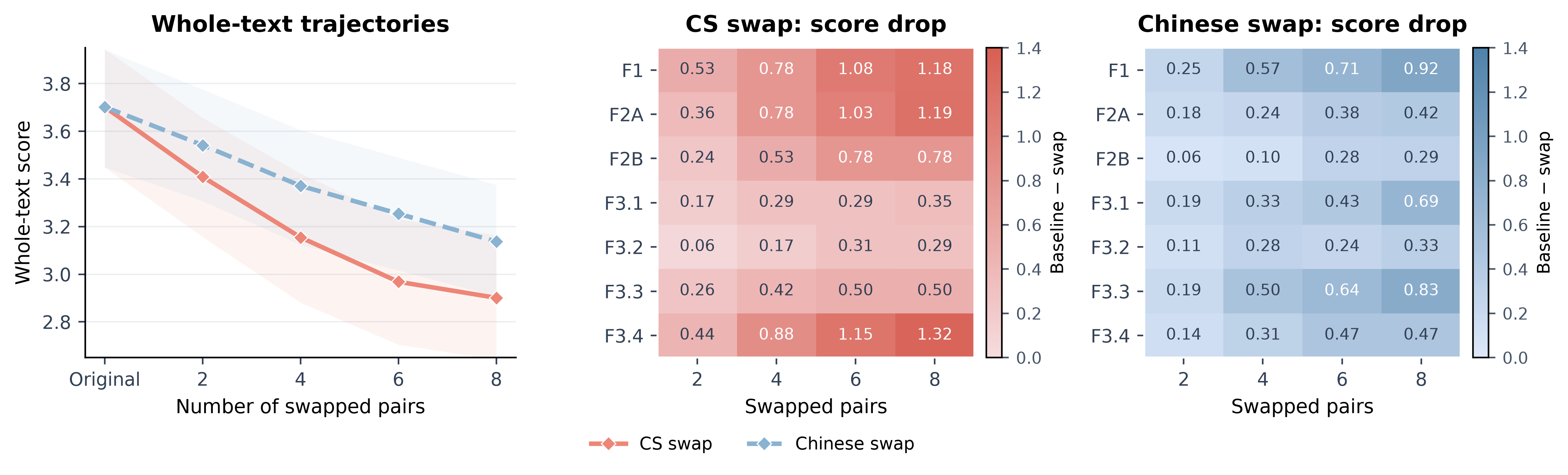}
\caption{Narrative salience and position-swap sensitivity. Top panels show mean Contextual Appropriateness (D2) and Functional Gain (D3) scores by salience level (S1--S3, increasing narrative consequence), with Q1--Q3 ranges and overlaid medians. Bottom panels show whole-text score changes as increasing numbers of code-switching event pairs or Chinese-span pairs are swapped, comparing sensitivity to the two types of disruption.}
\label{fig:construct-sensitivity}
\end{figure*}

\subsubsection{Construct Sensitivity}
\label{sec:construct-sensitivity}

Our perturbation and counterfactual studies test sensitivity to relevant evidence and controlled alternatives to observed language choices.

For Functional Gain, we conduct a cross-judge construct-sensitivity experiment on 72 events from 71 narratives, using LLM-generated Chinese-only counterfactuals reviewed and revised by human annotators. We use a reciprocal cross-judge design: one judge assigns events to HIGH and LOW groups based on its original D3 scores, while the other evaluates the original--counterfactual pairs in both presentation orders. We repeat the procedure with the judges' roles reversed. They determine whether the original CS realization better preserves contextually supported communicative and narrative effects, whether the alternatives are equivalent, or whether the monolingual alternative is stronger (Table~\ref{tab:functional-gain}). HIGH-D3 events more often favor the original CS realization, whereas LOW-D3 events yield more ties, consistent with greater replaceability by a natural monolingual alternative.

\begin{table}[ht]
\centering
\scriptsize
\setlength{\tabcolsep}{3pt}

\begin{tabular}{lcccc}
\toprule
Group & \makecell{Original CS\\stronger} & Tie & \makecell{Monolingual\\counterfactual stronger} & Directional score \\
\midrule
HIGH & 56.9\% & 20.8\% & 22.2\% & +0.347 \\
LOW & 29.9\% & 48.6\% & 21.5\% & +0.083 \\
\bottomrule
\end{tabular}
\caption{Functional Gain counterfactual judgments. The 72 events are divided into HIGH and LOW groups by original D3 scores. Original code-switching realizations are compared with human-reviewed Chinese-only alternatives through cross-judge comparisons in both presentation orders. Percentages report pooled outcomes, while directional scores assign $+1$ to original-CS wins, $0$ to ties, and $-1$ to monolingual wins.}
\label{tab:functional-gain}
\end{table}

Position-swap experiments on 36 Extended narratives test whole-text multilingual organization (Appendix~\ref{app:position-swaps}). We swap 2, 4, 6, and 8 \textbf{CS-event }pairs, alongside \textbf{Chinese-span} controls using the same schedule. Both manipulations increasingly reduce Whole-text scores: at the strongest level, scores fall from 3.702 to 2.901 under CS swaps and to 3.137 under Chinese swaps. CS swaps primarily disrupt multilingual organization; Chinese swaps have relatively greater effects on general narrative execution. This contrast indicates particular sensitivity of the multilingual-organization dimensions to CS-event placement.

\subsection{Multilingual Integration as a Major Axis of Model Differentiation}
\label{sec:multilingual-integration}

We compare cross-model separation across seven Extended Whole-text dimensions and their responses to prompt difficulty. Difficulty is estimated from total scores of six representative models using unweighted local D2/D3 means; the lowest- and highest-scoring prompt quartiles define hard and easy prompts. The remaining six models compare dimension-wise performance across these groups.

Among the seven Whole-text dimensions, M-F3.4\textbf{ (Multilingual Literary Integration)} shows the greatest variation across model means (SD = \textbf{0.742}). Multilingual-focused dimensions exhibit larger performance declines on harder prompts than general narrative dimensions. Further results and robustness analyses are reported in Appendix~\ref{app:uncertainty}.

\subsection{Narrative Salience and Event-Level Quality}
\label{sec:salience-quality}

We further examine how narrative salience relates to event-level code-switching quality within individual narratives (Figure~\ref{fig:construct-sensitivity}). Salience is positively associated with both Contextual Appropriateness and Functional Gain. We globally rank both variables within judge, center these ranks within articles, and average the Pearson correlations across the two cross-judge directions. The resulting correlations are $r=.642$ and $r=.496$ in Focused Narrative (4,884 articles), and $r=.768$ and $r=.789$ in Extended Narrative (4,800 articles), for Contextual Appropriateness and Functional Gain, respectively. Across both settings, higher narrative salience is associated with stronger contextual and functional realization, with larger correlations in Extended Narrative.

\subsection{From Local Code-Switching Quality to Whole-Text Performance}
\label{sec:local-to-whole}

In Extended Narrative, we average both cross-judge directions, pairing one judge's event-level scores with the other's Whole-text scores. Contextual Appropriateness and Functional Gain correlate with Whole-text performance at $r=.254$ and $r=.255$, respectively; individual directions appear in Table~\ref{tab:cross-judge}. These associations reduce influence from shared within-judge scoring tendencies.

Decomposing the Whole-text rubric further, all subdimensions remain positively associated with event-level CS quality. Dimensions more directly concerned with CS and whole-text multilingual organization show stronger associations than the general narrative dimensions, suggesting that event-level and Whole-text assessments capture related but distinct aspects of CS quality.

\section{Conclusion}

We introduce \textbf{CSWBench}, a benchmark for evaluating context-sensitive code-switching in creative writing, together with \textbf{CSWEval}, a theory-informed framework for evaluation at the event and whole-text levels. Experiments with twelve LLMs show that the main challenges of code-switching generation extend beyond producing locally plausible switches. Models differ substantially in how well they sustain multilingual choices as part of an evolving narrative system, while the quality of individual switches is positively associated with their narrative importance and shows moderate positive associations with whole-text performance. These results motivate evaluating code-switching as a local language choice and as a narrative pattern that develops across the text, with event-level and whole-text evaluation providing complementary and convergent evidence. We hope \textbf{CSWBench} and \textbf{CSWEval} support future research on multilingual generation and code-switching evaluation.

\section*{Limitations}

\paragraph{Language and text scope.}
CSWBench covers only Chinese-dominant narratives with Latin-script English
insertions. It does not address English-dominant text or other language pairs,
and the sociolinguistic conventions reflected in our sources may not transfer
to other multilingual communities. Target lengths of about 700 and 4,000
non-whitespace code points also leave long-form writing outside the
benchmark's scope.

\paragraph{Model and pipeline dependence.}
GPT-5.6-Sol is used in prompt construction, event annotation, and judging,
and is also one of the evaluated models. Excluding GPT-side scores for
GPT-family outputs leaves the model ordering unchanged (Appendix~\ref{app:self-preference}), but in
Focused Narrative the GPT-side judge still rates GPT-family outputs about
0.5 points (/20) more favorably relative to human ratings (Table~\ref{tab:self-family}). In a
separate 20-Seed sensitivity analysis, changing the event annotator shifts
Extended scores by a pooled mean of $+0.745$ points (/20), despite broadly
stable model rankings (Appendix~\ref{app:pipeline-sensitivity}).

\paragraph{Source data and character construction.}
About 82\% of the final prompts (661 of 807) derive from SwitchLingua, which
is synthetically generated, while the remaining ASCEND-derived prompts are
grounded in spontaneous conversation. Seeds are subsequently expanded with
model-generated narrative scaffolding, including participant profiles and
language backgrounds. The benchmark evaluates code-switching in creative-writing contexts constructed from bilingual source material. Generalization to naturally occurring interaction requires separate investigation.

\section*{Ethical Considerations}

CSWBench is designed to evaluate generated Chinese--English
creative writing, not the linguistic competence of real
bilingual speakers. Code-switching practices vary across
communities and contexts, and our constructed narratives
and LLM-based judgments may reflect limited sociolinguistic
norms or cultural stereotypes.

Our source-screening procedure excludes records flagged
for personal identifiers or sensitive content
(Appendix~\ref{app:construction}). The benchmark is not
intended for profiling or evaluating individual speakers.

\bibliography{references}

\appendix

\raggedbottom

\setcounter{topnumber}{4}
\setcounter{totalnumber}{6}
\setcounter{dblbotnumber}{0}
\renewcommand{\topfraction}{0.92}
\renewcommand{\textfraction}{0.06}
\renewcommand{\floatpagefraction}{0.62}
\makeatletter
\setlength{\@fptop}{0pt}
\setlength{\@fpsep}{14pt plus 1fil}
\setlength{\@fpbot}{0pt}
\setlength{\@dblfptop}{0pt}
\setlength{\@dblfpsep}{10pt}
\setlength{\@dblfpbot}{0pt plus 1fil}
\makeatother

\input{appendix_support/setup.tex}
\input{appendix_support/data_stats.tex}
\input{appendix_support/construction.tex}
\input{appendix_support/theory.tex}

\input{appendix_support/prompt_templates.tex}
\input{appendix_support/annotation.tex}
\input{appendix_support/scoring.tex}
\input{appendix_support/rubrics.tex}

\input{appendix_support/validation.tex}
\input{appendix_support/results.tex}

\input{appendix_support/analyses.tex}
\input{appendix_support/examples.tex}

\input{appendix_support/provenance.tex}

\end{document}

%% file: appendix_support/main_overall_performance.tex
\begin{table*}[!t]
\centering\small
\setlength{\tabcolsep}{3.4pt}
\renewcommand{\arraystretch}{1.05}
\resizebox{0.9\textwidth}{!}{%
\begin{tabular*}{\textwidth}{@{\extracolsep{\fill}}l|cccc c|cccc c@{}}
\toprule
\multirow{2}{*}{\textbf{Models}} & \multicolumn{5}{c|}{\textbf{Focused Narrative}} & \multicolumn{5}{c}{\textbf{Extended Narrative}} \\
\cmidrule(lr){2-6}\cmidrule(lr){7-11}
 & \textbf{D1} & \textbf{D2} & \textbf{D3} & \textbf{Whole} & $\Sigma$ & \textbf{D1} & \textbf{D2} & \textbf{D3} & \textbf{Whole} & $\Sigma$ \\
\midrule
\multicolumn{11}{@{}l}{\textit{Proprietary Models}} \\
\midrule
\addlinespace[1pt]
GPT-5.6-Sol & \underline{4.998} & \underline{3.349} & \underline{3.348} & \underline{3.806} & \underline{15.501} & 4.881 & \textbf{4.102} & \textbf{4.032} & \textbf{4.752} & \textbf{17.767} \\
GPT-5.6-Terra & 4.997 & 3.333 & 3.340 & 3.740 & 15.410 & 4.769 & \underline{4.069} & \underline{4.020} & 4.494 & \underline{17.352} \\
GPT-5.5 & \textbf{4.999} & \textbf{3.357} & \textbf{3.362} & \textbf{3.819} & \textbf{15.537} & 4.781 & 4.007 & 3.946 & \underline{4.614} & 17.348 \\
Gemini-3.5-Flash & 4.995 & 3.323 & 3.299 & 3.276 & 14.893 & \textbf{4.894} & 3.871 & 3.674 & 4.032 & 16.471 \\
Qwen3.7-Plus & 4.989 & 3.245 & 3.177 & 2.967 & 14.376 & 4.868 & 3.807 & 3.582 & 3.808 & 16.065 \\
Qwen3.7-Max & 4.990 & 3.291 & 3.245 & 3.014 & 14.540 & \underline{4.893} & 3.828 & 3.608 & 3.891 & 16.220 \\
\midrule
\multicolumn{11}{@{}l}{\textit{Open-weight Models}} \\
\midrule
\addlinespace[1pt]
DeepSeek-V4-Pro & 4.996 & 3.317 & 3.314 & 3.487 & 15.114 & 4.629 & 3.871 & 3.788 & 4.279 & 16.567 \\
DeepSeek-V4-Flash & 4.992 & 3.345 & 3.332 & 3.633 & 15.302 & 4.812 & 3.971 & 3.864 & 4.314 & 16.962 \\
MiniMax-M2.5 & 4.941 & 3.075 & 2.983 & 2.629 & 13.628 & 4.407 & 3.160 & 2.912 & 2.977 & 13.456 \\
MiniMax-M2.7 & 4.938 & 3.099 & 3.026 & 2.651 & 13.714 & 4.439 & 3.278 & 3.037 & 3.064 & 13.818 \\
Qwen3.5-397B-A17B & 4.992 & 3.237 & 3.171 & 2.958 & 14.357 & 4.639 & 3.660 & 3.481 & 3.710 & 15.491 \\
Qwen3.5-35B-A3B & 4.985 & 3.178 & 3.101 & 2.827 & 14.090 & 4.647 & 3.486 & 3.243 & 3.322 & 14.698 \\
\midrule
\textit{Average} & 4.984 & 3.262 & 3.225 & 3.234 & 14.705 & 4.722 & 3.759 & 3.599 & 3.938 & 16.018 \\
\bottomrule
\end{tabular*}
}
\caption{\textbf{Main Results of CSWBench.} Dual-judge mean scores for twelve LLMs on Focused and Extended Narrative. D1, D2, and D3 represent Licensing, Contextual Appropriateness, and Functional Gain, respectively; Whole denotes task-specific whole-text quality. Each dimension is scored out of 5, and $\Sigma$ is their sum (/20). Bold and underlined values indicate the highest and second-highest scores in each column, respectively.}
\label{tab:overall-performance}
\end{table*}

%% file: appendix_support/setup.tex
\section{Details of Experimental Setup}
\label{app:setup}

The model roster contains twelve models: GPT-5.6-Sol, GPT-5.6-Terra,
GPT-5.5, Gemini-3.5-Flash, Qwen3.7-Plus, Qwen3.7-Max, DeepSeek-V4-Pro,
DeepSeek-V4-Flash, MiniMax-M2.5, MiniMax-M2.7, Qwen3.5-397B-A17B, and
Qwen3.5-35B-A3B. The intended design is one narrative per model and prompt;
Appendix~\ref{app:results} reports observed coverage. Generation identities are
hidden during evaluation, and the two primary judges are GPT-5.6-Sol and
DeepSeek-V4-Flash.

\subsection{Generation Configuration}
All twelve models were called through the same OpenAI-compatible generation
interface. We explicitly set temperature to $0.7$, top-$p$ to $0.7$, and the
maximum output budget to 30,000 tokens for every model. Generation-side
reasoning was active for all twelve models under the model-specific
configurations listed in Appendix~\ref{app:provenance}. The four Qwen models
used the provider-default thinking mode, while the two MiniMax models
required explicit activation. For reproducibility, the released configuration
explicitly enables Qwen thinking on compatible API endpoints. Focused
and Extended narratives target approximately 700 and 4,000 non-whitespace
Unicode code points, respectively, which are the length bands stated in the
task templates. Each model receives the visible prompt of one Prompt Card and
produces one narrative.

\subsection{Scoring Configuration}
Judges use the predefined \textbf{CSWEval} rubrics. Both judges were queried at
high reasoning effort, which is separate from the generation-side reasoning
configuration recorded for each evaluated model in Appendix~\ref{app:provenance}.
GPT-5.6-Sol was used at the same high-reasoning configuration for Seed
construction, Prompt construction, event discovery, Extended evidence
construction, and quality judging.
Licensing (D1) is evaluated
event by event through the participant- and world-level licensing checks. For
each judge, article-level D1 is exactly $5$ times the proportion of valid events
that pass both checks. Within each task, the applicable dimension scores are
averaged before article-level reporting, and the
two judges' article-level scores are averaged before model-level aggregation.
Appendix~\ref{app:rubric} gives the exact aggregation and eligibility rules.

%% file: appendix_support/data_stats.tex
\section{Details of Dataset Statistics}
\label{app:data}

\subsection{Units and Source Normalization}
ASCEND supplies spontaneous spoken Chinese--English interactions~\cite{lovenia-etal-2022-ascend}, and SwitchLingua supplies controlled, synthetically generated text~\cite{xie2025switchlingua}; both are normalized into a single record pool. A source row, conversational turn, window, Seed, and task Prompt are different units.

\begin{table*}[t]
\centering\small
\setlength{\tabcolsep}{7pt}

\begin{tabular}{lrrr}
\toprule
Stage & Total & ASCEND & SwitchLingua \\
\midrule
Source records sampled & 3,500 & 382 & 3,118 \\
Retained source records & 3,009 & 287 & 2,722 \\
Validated Seeds retained for benchmark construction & 596 & 109 & 487 \\
Final evaluation prompts & 807 & 146 & 661 \\
\quad Focused & 407 & 76 & 331 \\
\quad Extended & 400 & 70 & 330 \\
\bottomrule
\end{tabular}
\caption{Construction and benchmark-selection statistics.}
\label{tab:data-funnel}
\end{table*}

\subsection{Final-Panel Coverage Audit}
The final 807 prompts are linked to 546 distinct Seeds from the pool of 596 validated Seeds. The two tasks share 261 Seeds (Jaccard overlap .478), while Focused has 146 task-specific Seeds and Extended has 139.

\begin{table*}[t]
\centering\small

\resizebox{\textwidth}{!}{%
\begin{tabular}{lrrr}
\toprule
Coverage item & Focused ($n=407$) & Extended ($n=400$) & Cross-task \\
\midrule
Scene domains & 13/14 & 14/14 & Largest gap 5.0 pp \\
Participant relations & 6/9 & 8/9 & Largest gap 4.8 pp \\
Interaction patterns & 11/11 & 11/11 & All retained \\
Preferred styles & 5/5 & 5/5 & All baseline styles retained \\
Unique source groups & 350 & 340 & 88 / 93 prompts repeat a group \\
Stable scene/relation/style & 12 / 5 / 5 & 14 / 8 / 5 & $\geq$ two source groups \\
\bottomrule
\end{tabular}
}
\caption{Descriptive coverage of the Focused and Extended prompt sets, including the largest cross-task prevalence gap for each category. A source group is an ASCEND conversation or a SwitchLingua record to which a prompt's Seed is bound; a category is stable when covered by at least two distinct source groups.}
\label{tab:panel-coverage}
\end{table*}

\begin{figure*}[t]
\centering
\includegraphics[width=\textwidth]{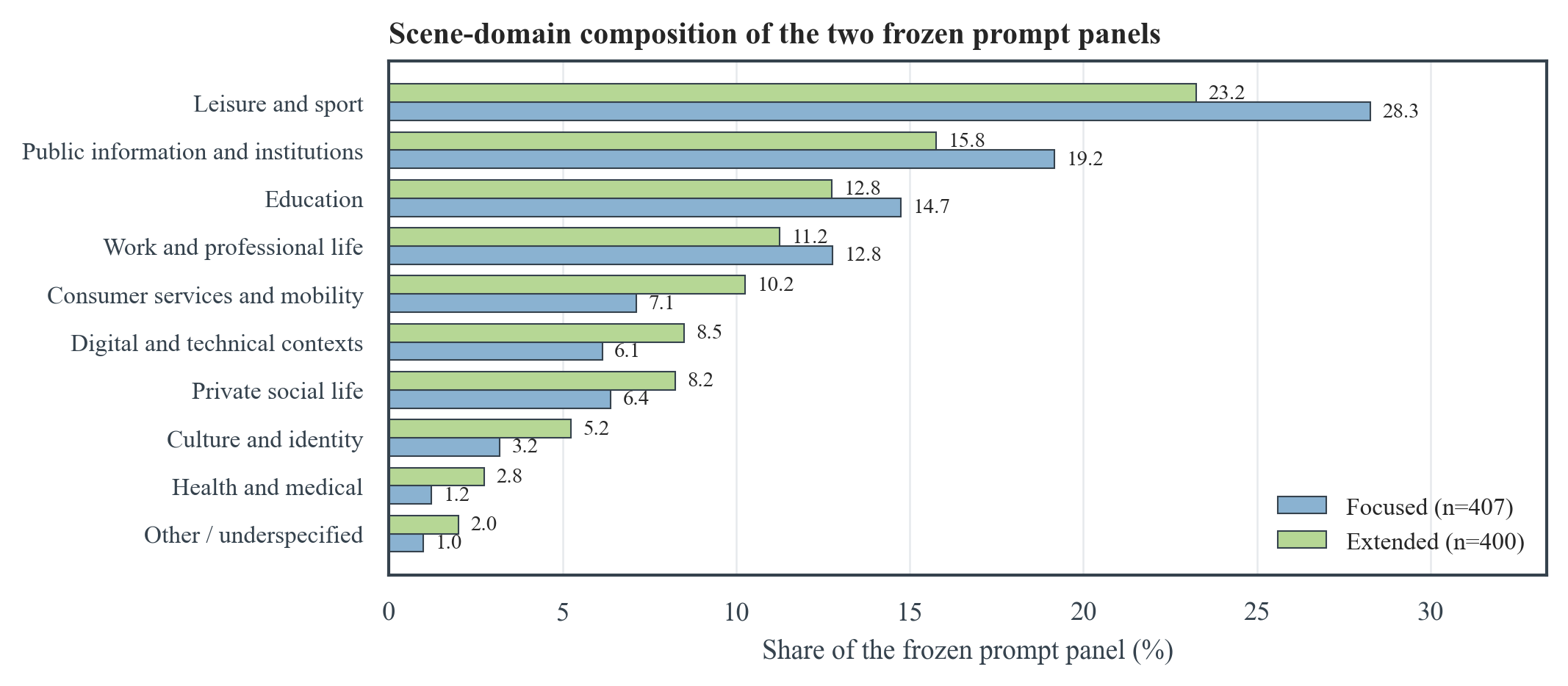}
\caption{Scene-domain distribution of the two prompt sets over ten reporting categories, which aggregate the 14 recorded scene domains, for prompts derived from ASCEND and SwitchLingua~\cite{lovenia-etal-2022-ascend,xie2025switchlingua}. The ten categories are consumer services and mobility, culture and identity, digital and technical contexts, education, health and medical, leisure and sport, other or underspecified, private social life, public information and institutions, and work and professional life.}
\label{fig:coverage-scene}
\end{figure*}

\begin{figure}[t]
\centering
\includegraphics[width=\columnwidth]{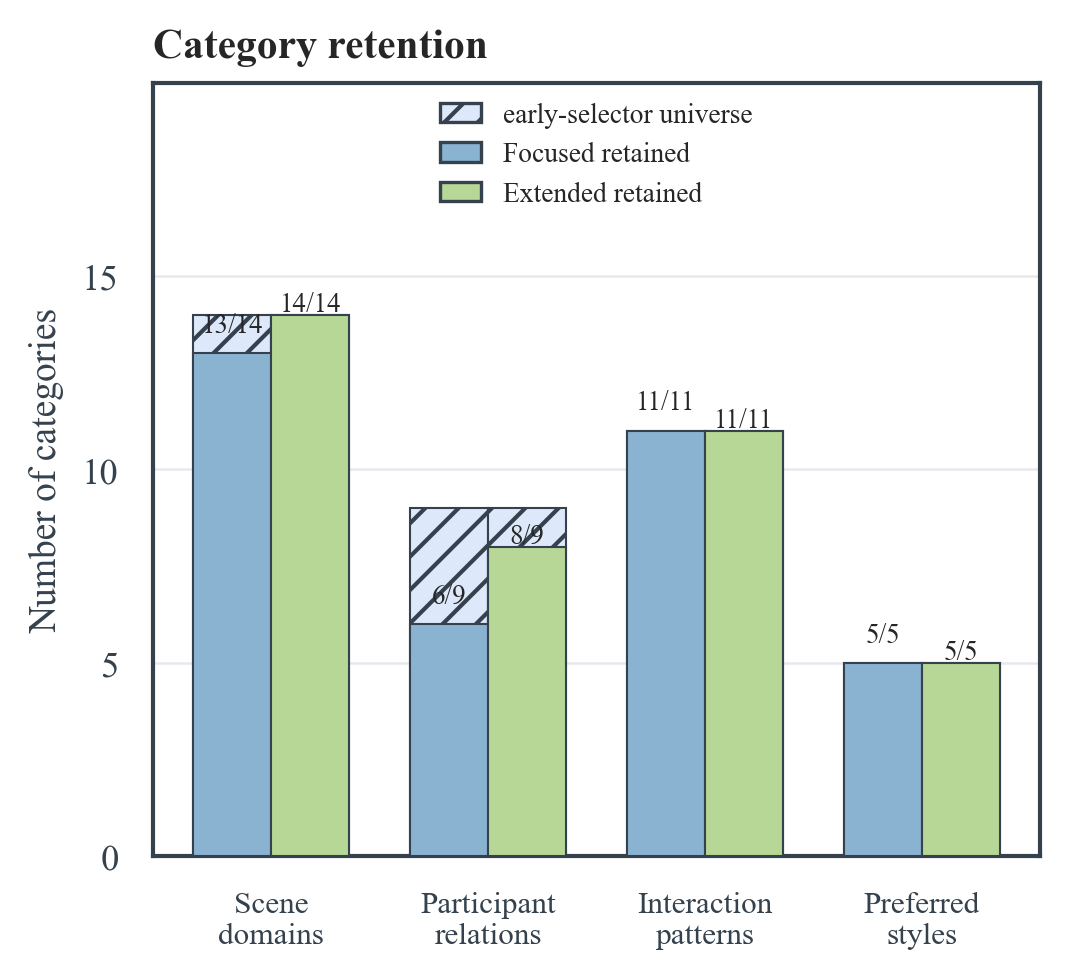}
\caption{Coverage of the recorded category universes. Each bar pair gives the number of categories retained by the Focused and Extended sets against the recorded universe size for that coverage item.}
\label{fig:coverage-retention}
\end{figure}

The Focused set omits one recorded scene domain and three relation categories; Extended omits one relation category. Both sets retain all recorded interaction-pattern and style categories. Table~\ref{tab:panel-coverage} reports what each set retains, together with the largest cross-task prevalence gap for each item. Figure~\ref{fig:coverage-scene} gives the scene-domain distribution, and Figure~\ref{fig:coverage-retention} the retained-versus-recorded counts.

%% file: appendix_support/construction.tex
\section{Data and Seed Construction Details}
\label{app:construction}

\subsection{Operational Source Screening}
The source-screening prompt first checks P0 (interpretable coherent discourse), then P1 (attributable semantic CS), followed by sensitivity flags. A retained record needs enough context to identify a coherent interaction or discourse unit and at least one meaningful English use by a current speaker/voice. Ordinary oral hesitation is not corruption. English words alone do not establish CS.

\begin{itemize}
\item \textbf{Structural exclusions:} corrupt/truncated material, fragments without recoverable context, unrelated units concatenated together, and detached word lists. Co-occurrence of separate monolingual speakers or unrelated units is insufficient without attributable switching.
\item \textbf{Lexical borrowing and names:} conventionalized borrowings, names, brands, or labels alone do not demonstrate a switch. The source reason categories include \texttt{BORROWING\_ONLY} and \texttt{NAMED\_ENTITY\_ONLY}.
\item \textbf{Quotation and mediated material:} quoted English, copied material, or static display without active uptake by the current discourse voice is excluded (\texttt{QUOTATION\_ONLY}, \texttt{MEDIATED\_ONLY}). Active reuse can qualify; quotation marks alone are not the criterion.
\item \textbf{Other non-CS cases:} cross-unit language separation, noisy language detection, English-dominant material with isolated Chinese, or unresolved attribution receive \texttt{NOT\_TRUE\_CSW} or \texttt{AMBIGUOUS}, rather than being forced into the retained set.
\item \textbf{Sensitivity:} flagged personal identifiers/privacy, minors-related risk, sexual content, self-harm, severe violence, illegal conduct, hate, or private health information are excluded under the screening prompt. Mere ordinary mention of a topic is not automatically a sensitive case.
\end{itemize}

Across the combined pool of 3,500 source records, 3,401 records pass P0 and 3,309 satisfy P1; after the subsequent sensitivity screen, 3,009 records are retained for Seed construction.

\subsection{Seed Construction and Validation}
The source must support the topic, evidence summary, source-supported possible situation, and identifiable interactional/medium cues. Narrative pressure, a plausible interaction sequence, relation detail, stylistic affordances, and open consequences can be added \emph{only as explicitly labeled creative scaffolding}. Seed representation is described in Appendix~\ref{app:seed-representation}; a stored Seed and the construction examples are reproduced in Appendix~\ref{app:examples}.

Seed-stage validation is performed by DeepSeek-V4-Pro using two checks. Gate A verifies that material evidence claims are supported by the redacted source, and Gate B verifies that labeled creative scaffolding remains consistent with the source basis and constraints. A Seed is eligible only when both checks pass.

The task-specific prompt sets are selected from the pool of 596 validated Seeds, prioritizing coverage of scene domain, participant relation, interaction pattern, and preferred style, with deterministic tie-breaking and source-group constraints. For records with available source English-token counts, construction refusal rates across the four source-CS length quartiles range from 45.2\% to 54.1\%. After Prompt construction, every completed Prompt undergoes independent human screening for grounding fidelity and scaffolding compatibility (Appendix~\ref{app:validation}).

\subsection{Source Licenses and Release Terms}
\label{app:source-licenses}

ASCEND~\cite{lovenia-etal-2022-ascend} is distributed
under CC BY-SA 4.0, while
SwitchLingua~\cite{xie2025switchlingua} is distributed
under CC BY-NC 4.0. Both datasets are used for
non-commercial academic research, and their
original records are not redistributed.

Original CSWBench Prompt Card contributions are
released under CC BY-NC 4.0, and the accompanying
code under Apache-2.0.

%% file: appendix_support/theory.tex
\section{Theory Review and Construct Development}
\label{app:theory}

\subsection{Search and Selection Trail}
The theory review uses two routes: theory-led citation tracing from anchor works, and a structured bibliographic search. The executed OpenAlex log contains 32 title-and-abstract queries , with publication dates restricted to 1950--2026-06-01, pagination at 200 records, and no initial language or publication-type restriction. It records 1,972 query occurrences, 1,615 distinct OpenAlex IDs, and 1,518 works after version deduplication.

The executed searches are organized around the two theoretical foundations used in the main text. For interactional approaches to code-switching, the registered title-and-abstract queries include \textit{``code-switching'' AND interactional AND sequential}, \textit{``code-switching'' AND ``contextualization cue''}, \textit{``code-switching'' AND ``audience design''}. For literary multilingualism, they include \textit{``literary code-switching'' AND narrative}, \textit{``literary multilingualism'' AND narrative AND function}, \textit{``staged code-switching''}, and \textit{``fictional multilingualism''}. The complete master log contains 162 completed entries: 32 structured OpenAlex searches, 108 OpenAlex anchor-resolution/backward/forward-chain requests, and 22 targeted scholarly-web searches with direct publisher or official-source follow-up. Google Scholar and arXiv were used for supplementary discovery and computational-comparator location.

The citation-tracing protocol starts from representative theoretical anchors, checks their references for predecessors, and follows citing work for extensions, critiques, and boundary conditions. It records parent source and tracing direction. Its planned stopping rule is two rounds, with an optional third round for unresolved conceptual gaps. Deduplication prioritizes DOI, then bibliographic ID, then normalized title, author, and year, while linking versions.

\paragraph{Eligibility.}
Substantive theory records must contribute to one of the two theoretical foundations used in the main text: (i) interactional approaches to code-switching, which treat language choice as situated social action shaped by participants, local interactional conditions, and unfolding discourse; or (ii) literary multilingualism, which treats multilingual language in fiction as a deliberately shaped literary and narrative resource. These records must expose an evaluative distinction and identify textual evidence that could support judging it in generated narratives. Methodological comparators are retained when they inform review methodology, construct operationalization, or the use of textual evidence in evaluation. Books and chapters are eligible alongside articles. Duplicate versions, unverifiable or non-academic items, and claims requiring inaccessible real-world cognition are excluded.

\paragraph{Source count.}
The final convention is \textbf{107 retained sources $\rightarrow$ 70 concepts $\rightarrow$ six preliminary constructs $\rightarrow$ four quality constructs}. The 107-record union contains 63 theory-led and 44 search-led records, comprising 100 substantive theory records and seven methodological comparators.

\subsection{From Concepts to Constructs}
The 70 candidate concepts comprise 44 theory-led and 26 search-led formulations.
Each records an evaluative question together with the evidence conditions and scope
that would settle it. Conceptual consolidation yields six preliminary constructs,
of which four remain quality constructs.

Consolidation compares the judgment object, evidence window, and failure condition
of each concept, grouping formulations where these agree. Table~\ref{tab:construct-mapping}
gives the primary assignments.

\begin{table*}[t]
\centering\small

\begin{tabular}{p{.27\textwidth}p{.30\textwidth}rp{.26\textwidth}}
\toprule
Candidate concept families & Preliminary construct & Count & Final role \\
\midrule
Repertoire, access, represented resources & Participant--world resource grounding & 9 & D1 Licensing \\
\cmidrule(lr){1-4}
Voice, recipient, sequence, situation and local fit & Situated interactional coherence & 22 & D2 Contextual Appropriateness \\
\cmidrule(lr){1-4}
Contrastive meaning, stance, relationship, style and experience & Functional multilingual contribution & 19 & D3 Functional Gain \\
\cmidrule(lr){1-4}
Recurrence, contrast, development and literary integration & Whole-text multilingual orchestration & 4 & Extended Whole-text; Focused uses general literary/narrative quality \\
\cmidrule(lr){1-4}
Textual warrant, mediation and inferential boundaries & Evidentially constrained literary mediation & 9 & Cross-cutting evidence constraint, not a fifth quality score \\
\cmidrule(lr){1-4}
Eventfulness, consequence and narrative importance & Narrative event importance & 7 & Salience S1--S3 for D2/D3 weighting, not quality \\
\bottomrule
\end{tabular}
\caption{Condensed concept-to-construct mapping. Counts sum to 70 primary assignments.}
\label{tab:construct-mapping}
\end{table*}

The two remaining constructs inform evidence requirements and salience weighting. The four retained quality constructs distinguish permission, local fit, functional contribution, and whole-text organization.

%% file: appendix_support/prompt_templates.tex
\section{Prompt Templates and Representative Prompt Cards}
\label{app:prompt}

\subsection{Actual Seed Representation}
\label{app:seed-representation}
A Seed is a structured record. It keeps a
source-grounded layer and an explicitly added scaffolding layer separate, so that
constructed narrative material can never be reread as source fact. One authentic Seed, with its two layers presented separately, is provided in Appendix~\ref{app:examples}.

\subsection{Seed-Construction Instruction}
Seed construction is an LLM stage that reads a screened record and returns either
a usable, explicitly bounded enrichment proposal or a documented refusal. Representative Seed-construction instruction excerpts appear in
Appendix~\ref{app:examples}.

\subsection{Task Templates and Representative Samples}
Each task template is the four-paragraph form instantiated by the representative
Prompt Cards: paragraphs that stay fixed across prompts, plus the Seed-conditioned
paragraphs rendered from the fields of the Seed. Two stored Prompt Cards, one
Focused and one Extended, are reproduced in Appendix~\ref{app:examples};
both are built from the \emph{same} Seed, which makes the two task mappings
directly comparable.

\subsection{Generation Message Envelope}
For Focused Narrative, a fixed system message preceded the four-paragraph
Prompt Card sent as the user message. It required only narrative-body output:
no title, explanation, analysis, counting procedure, Markdown fences, or
other non-narrative content. For Extended Narrative, the four-paragraph
Prompt Card was the sole user message.

%% file: appendix_support/annotation.tex
\section{Annotation and Output Contracts}
\label{app:annotation}

\subsection{LLM Event Discovery and Boundary Fixing}
GPT-5.6-Sol serves
as the discoverer in both settings, and the quality judges score the resulting
event manifest. What
follows is the operational definition the discovery stage applies.

A semantic event is a locally coherent English choice attributable to a speaker
or narrative voice and to an ongoing discourse action. It can consist of a word,
a phrase, a clause, or a sentence, and it may cover several mechanically
extracted Latin-script spans. Continuous English with the same voice and the same
discourse action belongs to one event. The discoverer
classifies the candidate spans, proposes the semantic groupings, and never equates
the number of mechanical spans with the number of events. Accepted proposals are
fixed before scoring, and the quality judges cannot merge, split, add, or reorder
that manifest. Independent human validation of both candidate inclusion and event
grouping is reported in Appendix~\ref{app:human}.

\paragraph{Inclusion and exclusion.}
Discovery asks whether the English functions as a current, attributable language
choice. Lexical borrowings, proper names, quotations, and detached or mediated multilingual
material are not CS solely because Latin letters occur, whereas active uptake or
reuse by the current speaker can qualify when it performs an interactional or
narrative action. Event inclusion is independent of downstream quality: a genuine
code-switching event may still be contextually inappropriate, poorly motivated, or
functionally weak, which is assessed subsequently by D2 and D3.

Examples of discovery instructions implementing these rules appear in Appendix~\ref{app:examples}.

\subsection{Longitudinal Evidence Construction}

For Extended Narrative, a separate evidence-construction stage records character and language resources, changing character and relationship states, narrative-world facts, non-switching behavior, voice, transitions, and ending conditions. GPT-5.6-Sol constructs these evidence records before quality judging. Both judges receive the same records, which remain unchanged during scoring.

For event-level Licensing, admissible evidence citations are restricted to the visible Prompt and article evidence units ending before the corresponding code-switching event begins. Across all 4,800 Extended narratives, the resulting D1 evidence allowlists contain no future-evidence violations. Both judges receive the full narrative and frozen evidence records for joint event-level and whole-text scoring. We separately evaluate the factual and temporal fidelity of these evidence records in Appendix~\ref{app:evidence-fidelity}.

%% file: appendix_support/scoring.tex
\section{Scoring Instructions}
\label{app:scoring}

The quality judgments reported in this paper were produced by two primary judges
under the quality-judging instructions illustrated in Appendix~\ref{app:examples}. In both
settings the events are already finalized before scoring: the judge never decides
what counts as a code-switching event, and it never rewrites the narrative. The
two settings differ in how the event-level dimensions are obtained. In Focused
Narrative the judge scores the applicable subdimensions separately and the
article-level dimension score is formed by aggregating them; in Extended Narrative
the same criteria serve as guidance and the judge assigns one direct score per
dimension. Appendix~\ref{app:rubric} provides the dimension-specific and item-level anchors and the exact
aggregation rules these judgments feed into.

%% file: appendix_support/rubrics.tex
\section{\textbf{CSWEval} Rubrics and Aggregation}
\label{app:rubric}

This appendix specifies the paper's reporting rules. The scoring prompts supply the judgments; aggregation supplies article and model scores. 

\subsection{Licensing (D1)}
\paragraph{Two event-level decisions.}
Participant/role-level licensing asks whether the attributed speaker or narrative voice has the identity, language history, comprehension/expression resources, and established profile needed for this particular use of English. World-level licensing asks whether the addressee, carrier (speech, writing, inner speech, or active mediated reuse), scene, relationship state, social rules, established facts, and prior discourse state permit the choice. Each check receives \texttt{PASS} or \texttt{FAIL}; an event passes only if both pass.

For article $a$ and judge $j$, let $V_{aj}$ contain the semantic CS events with an interpretable \texttt{PASS}/\texttt{FAIL} decision on \emph{both} checks. Then
\begin{equation}
D1_{aj}=\frac{5}{|V_{aj}|}\sum_{e\in V_{aj}}p_{aje},
\end{equation}
where $p_{aje}=1$ exactly when both role and world checks pass, and is zero otherwise.
The current primary scoring schemas require binary decisions. D1 is \emph{not} salience-weighted.

\subsection{Event-Level Ordinal Anchors}
The event-level ordinal scale progresses from \textbf{1}, unsuccessful or clearly unsupported; \textbf{2}, weak or partial; \textbf{3}, clear and adequate; \textbf{4}, strong and specific; to \textbf{5}, precise and highly realized. The operational meaning of these levels is specified separately for Contextual Appropriateness and Functional Gain below.

\subsection{Contextual Appropriateness (D2)}
Focused D2 scores eight subdimensions at each event. Applicability is checked before quality, as specified in the table caption. Contextual Appropriateness evaluates how well the multilingual choice fits its local narrative and discourse context. Its 1--5 anchors are: \textbf{1}, contextually unsupported or conflicting; \textbf{2}, weak or partial contextual fit; \textbf{3}, clear contextual fit supported by concrete local evidence; \textbf{4}, strong and specific contextual fit; \textbf{5}, precise contextual realization, for which the particular wording, form, and placement are well supported by the context where relevant.

\begin{table*}[t]
\centering\scriptsize
\setlength{\tabcolsep}{3pt}

\begin{tabular}{p{.16\textwidth}p{.17\textwidth}p{.19\textwidth}p{.20\textwidth}p{.19\textwidth}}
\toprule
Item & 1 & 3 & 4 & 5 \\
\midrule
D2.1 Voice attribution & Wrong or indeterminate voice & Dialogue, thought, or narration clearly attributed & Distinguishes potentially confusable layered voices or quotation responsibility & Attribution remains precise across layered voices or quotation boundaries \\
\cmidrule(lr){1-5}
D2.2 Carrier fit & Expression incompatible with its carrier & Natural fit to speech, thought, narration, message, or reuse & Carrier and expression are specifically matched in form and use & Wording and form are precisely matched to the carrier and its constraints \\
\cmidrule(lr){1-5}
D2.3 Addressee trigger & Known addressee conditions conflict with the choice & Recipient's ability, identity, or response motivates it & Precisely responds to recipient's resources, wording, identity, or immediate uptake & The realization precisely fits the recipient's resources, wording, identity, and immediate interactional state \\
\cmidrule(lr){1-5}
D2.4 Topic trigger & Unrelated to current concept, object, or material & Topic directly supports English & Conceptual boundaries, source wording, or reference shape the local interpretation & Wording and placement are tightly grounded in the current concept, source wording, referent, or topic structure \\
\cmidrule(lr){1-5}
D2.5 Emotion trigger & Applicable emotional mechanism cannot be located & Switch accompanies emotional change & Occurs at a specific emotional threshold, defensive move, or intimacy shift & Wording and placement precisely align with the established emotional state or transition \\
\cmidrule(lr){1-5}
D2.6 Power, stance, identity & Applicable negotiation mechanism cannot be located & Clearly marks stance or identity & Changes the interaction at an assertion, concession, boundary-setting, alliance, or repositioning move & Wording and form precisely fit the current stance, power, or identity configuration \\
\cmidrule(lr){1-5}
D2.7 Register trigger & Required register change is not realized & Clear conventional register change & Register contrast works with the relationship, occasion, and ongoing action & Wording and form precisely match the relationship, occasion, and interactional style \\
\cmidrule(lr){1-5}
D2.8 Local switch point & Syntax, meaning, reference, or turn continuity breaks & Natural and coherent boundary & Uses a particular syntactic boundary, rhythmic break, or turn transition & Wording and placement are precisely matched to the local syntax, rhythm, or turn structure \\
\bottomrule
\end{tabular}
\caption{Focused D2 rubric. NA is allowed for D2.3--D2.7; D2.1, D2.2, and D2.8 always require numeric scores. For D2.7, unchanged adjacent register is NA, whereas a required but unrealized transition is scored.}
\label{tab:d2-rubric}
\end{table*}

Extended D2 instead receives one direct event score. Voice, carrier, addressee, topic, emotion, power/stance, register, and switch boundary are \emph{guidance}, not separately scored fields, and the D2 anchors above are applied holistically across the relevant evidence.

\subsection{Functional Gain (D3)}
Functional Gain measures the added value that a code-switching choice itself provides. Credit follows the choice and its local effect rather than the rest of the sentence or scene. Its 1--5 anchors are: \textbf{1}, the relevant function is absent or poorly realized; \textbf{2}, weak, limited, or largely redundant functional value; \textbf{3}, a clear, observable, and locatable contribution within the current event; \textbf{4}, a strong and specific contribution that actively does work in the event; \textbf{5}, a substantial, precise, and non-redundant contribution that is tightly integrated into the current event.

Focused D3 scores the five subdimensions in Table~\ref{tab:d3-rubric} separately at each event and aggregates them into the dimension score.

\begin{table*}[t]
\centering\scriptsize
\setlength{\tabcolsep}{3pt}
\captionsetup{skip=4pt}

\begin{tabular}{p{.16\textwidth}p{.17\textwidth}p{.19\textwidth}p{.20\textwidth}p{.19\textwidth}}
\toprule
Item & 1 & 3 & 4 & 5 \\
\midrule
D3.1 Semantic precision & No added distinction, source reference, ambiguity control, or precision & Recognizable terminological/conceptual precision & Specific meaning, original wording, ambiguity, or wordplay affects understanding & The semantic contribution is exceptionally precise and tightly integrated into the event's local meaning \\
\cmidrule(lr){1-5}
D3.2 Relationship/stance gain & Language choice adds no relational or attitudinal signal & Clear relational or attitudinal signal carried by the language choice & Substantially strengthens a relational act or subtext & Carries a highly specific and strongly realized relational or attitudinal function within the interaction \\
\cmidrule(lr){1-5}
D3.3 Identity/cultural specificity & Merely labels an identity & Adds a plausible, specific identity connection & Closely connects experience and current self-positioning & Makes an exceptionally specific and well-supported contribution to identity or cultural positioning \\
\cmidrule(lr){1-5}
D3.4 Stylistic tension & Only typographic novelty or display & Clear local voice, rhythm, double-voicing, or restraint & Works with syntax, rhythm, imagery, or subtext to organize the passage & Is tightly integrated with local form to produce a highly realized stylistic effect \\
\cmidrule(lr){1-5}
D3.5 Experiential texture & Merely labels a background & Connects to lived experience and adds texture & Interacts closely with action, medium, institution, body, or memory & Makes the represented experience exceptionally specific, credible, and texturally realized within the event \\
\bottomrule
\end{tabular}
\caption{Focused D3 rubric. NA applies to D3.1, D3.2, D3.3, and D3.5 when the corresponding function is not applicable; D3.4 always requires a numeric score.}
\label{tab:d3-rubric}
\end{table*}

Extended D3 is a direct 1--5 event score. The same semantic, relational, identity-cultural, stylistic, and experiential checks serve as guidance rather than separately scored fields, and the D3 anchors above are applied holistically across the relevant evidence. Table~\ref{tab:functional-gain} and Figure~\ref{fig:functional-gain} report the separate counterfactual study that tests this construct in both settings.

\subsection{Focused Whole-Text Quality}
Focused Whole assesses \emph{overall literary and narrative quality}, not Multilingual Orchestration. L1--L5 are five scored dimensions without separately scored children. They require integers 1--5, with no NA/IE.

\begin{table*}[t]
\centering\scriptsize
\setlength{\tabcolsep}{3pt}
\captionsetup{skip=4pt}

\begin{tabular}{p{.16\textwidth}p{.18\textwidth}p{.19\textwidth}p{.19\textwidth}p{.19\textwidth}}
\toprule
Dimension & 1 & 3 & 4 & 5 \\
\midrule
L1 Completeness/coherence & Main line, causality, or ending not intelligible & Complete story with adequate causality and ending & Opening pressure, change, decisions, and ending form a sustained causal chain & Several indispensable nodes form a causal system \\
\cmidrule(lr){1-5}
L2 Voice, relationships, characterization & Indistinct characters, unsupported motives, or announced change & Distinct stances, credible motives, relationship advanced through interaction & Voice, action, and subtext differentiate characters over stages & Layered motives, avoidance, behavior, and decisions form a complex coherent system \\
\cmidrule(lr){1-5}
L3 Style, rhythm, imagery & Reading obstructed or basic unity absent & Fluent narration with local rhythmic/imagistic organization & A formal mechanism persists across locations & Voice, rhythm, imagery, and perspective form a distinctive system \\
\cmidrule(lr){1-5}
L4 Emotional/thematic depth & Implausible emotion or abstract conclusions replace narration & Plot, action, and dialogue support emotion/theme & Two mutually constraining values develop persistently & Layered conflicts revise each other and sustain interpretive tension \\
\cmidrule(lr){1-5}
L5 Originality/nonformulaic realization & Predominantly generic template & Some effective choices within a familiar whole & Mutually supporting non-obvious choices on at least two levels & Distinctive unified narrative principle constrained by earlier development \\
\bottomrule
\end{tabular}
\caption{Focused whole-text rubric. Score 2 denotes partial success between anchors 1 and 3; all five dimensions require numeric scores, with no NA/IE.}
\label{tab:focused-whole-rubric}
\end{table*}

\subsection{Extended Whole-Text Dimensions}
Seven dimensions are scored independently (Table~\ref{tab:extended-whole-rubric}). M-F2A and M-F2B are separate; asymmetric language use is permitted.

\begin{table*}[t]
\centering\scriptsize
\setlength{\tabcolsep}{3pt}

\begin{tabular}{p{.10\textwidth}p{.29\textwidth}p{.53\textwidth}}
\toprule
ID & Dimension & Evidence \\
\midrule
M-F1 & Multilingual narrative-system stability & Voice ownership; speech/thought/quotation boundaries; carriers; focalization; preservation/translation policy; language facts and rules maintained or coherently developed across scenes. \\
\cmidrule(lr){1-3}
M-F2A & Character A bilingual profile and language arc & Repertoire, function/relationship/affect mappings, scene responsiveness, motivated transitions and non-switching; the same multilingual person developing on textual evidence. \\
\cmidrule(lr){1-3}
M-F2B & Character B bilingual profile and language arc & The same checks independently for B, without requiring symmetry or rewarding event count alone. \\
\cmidrule(lr){1-3}
M-F3.1 & Structure, causality, pacing, and closure & Scene progression, causal continuity, escalation, turning points, proportional pacing, and an ending reflecting actions and consequences. \\
\cmidrule(lr){1-3}
M-F3.2 & Character and relationship development & Goals, knowledge, emotions, actions, and transitions supporting character development; relationship change earned through prior interaction. \\
\cmidrule(lr){1-3}
M-F3.3 & Scene, dialogue, prose, and emotional execution & Scene realization, distinct dialogue, clarity/control, sensory and atmospheric specificity, credible emotion, and engagement without ornament as a proxy. \\
\cmidrule(lr){1-3}
M-F3.4 & Multilingual Literary Integration & Sustained character-voice differentiation, sociocultural specificity, stylistic identity, and meaningful relations among social voices across the narrative. \\
\bottomrule
\end{tabular}
\caption{Extended whole-text rubric.}
\label{tab:extended-whole-rubric}
\end{table*}

The seven Extended whole-text dimensions use a separate whole-text scale: \textbf{1}, seriously unsupported, unstable, ineffective, or damaging; \textbf{2}, substantial weaknesses with limited successful evidence; \textbf{3}, adequate and generally supported, with meaningful limitations; \textbf{4}, strong, sustained, and well-supported execution; \textbf{5}, exceptional, precise, sustained, and difficult to improve materially. A score of 5 requires repeated or structurally important evidence rather than one attractive sentence.

\subsection{Narrative Salience}
Each quality judge assigns its own salience label to each event: S1 (weight 1) has little or no consequence beyond the immediate moment; S2 (weight 2) introduces a clear change with consequences for the current scene or subsequent narrative development; S3 (weight 3) introduces a major change whose consequences persist or substantially shape later narrative development. Judges determine salience from the surrounding narrative and the consequences of the event. Salience weights \emph{only D2 and D3}.

\subsection{Exact Aggregation and Eligibility}
For Focused dimension $d\in\{D2,D3\}$, let $A_{ajed}$ be the applicable numeric subdimensions for event $e$. Compute
\begingroup
\setlength{\abovedisplayskip}{4pt}
\setlength{\abovedisplayshortskip}{4pt}
\setlength{\belowdisplayskip}{4pt}
\setlength{\belowdisplayshortskip}{4pt}
\setlength{\jot}{2pt}
\begin{align}
q_{ajed}&=\frac{1}{|A_{ajed}|}\sum_{k\in A_{ajed}}x_{ajedk},\\
D^{(d)}_{aj}&=\frac{\sum_{e\in N_{ajd}}s_{aje}q_{ajed}}{\sum_{e\in N_{ajd}}s_{aje}},
\end{align}
\endgroup
where $N_{ajd}$ contains events with a numeric event mean and valid salience. Extended uses the same second equation with $q_{ajed}$ equal to the judge's \emph{direct} dimension score. NA/IE are omitted only at their corresponding level; they are not zeroes, and removing a subdimension does not remove the other valid scores of that event.

Whole-text dimensions are averaged separately: $Whole_{aj}=K^{-1}\sum_{k=1}^{K}W_{ajk}$, with $K=5$ for Focused and $K=7$ for Extended. All required whole scores must be present, and each article dimension is then averaged across the two primary judges: $D_a=(D_{a,1}+D_{a,2})/2$. An article is eligible when every whole-text dimension is present and each scored Focused dimension has at least one event with a numeric event mean and valid salience. The within-setting composite is $\Sigma_a=D1_a+D2_a+D3_a+Whole_a$ (/20), and the model-level composite is the mean over eligible articles.

%% file: appendix_support/validation.tex
\section{Prompt Validation and Human Evaluation}
\label{app:validation}

\subsection{Criteria and Decision Rules}
The written Focused expert-review form supplies eight checks: Q1 grounding fidelity, preserving grounded information without distortion; Q2 scaffolding compatibility, requiring added characters, relations, and scene details to remain plausible and consistent with that grounding; Q3 role/world permission and information access; Q4 observable interaction and consequential uptake; Q5 identifiable semantic CS without prescribing its realization; Q6 literary affordances and creative openness; Q7 compatible joint task constraints; and Q8 non-leading, complete visible instructions.

Each content-validity item uses 1--5: 1 fails the construct or is structurally unsuitable; 2 has a major defect; 3 has a substantive borderline ambiguity; 4 is adequate without a mandatory change; 5 is robust across reasonable realizations. These \emph{prompt screening} thresholds are separate from the evaluation score anchors in Appendix~\ref{app:rubric}. A paper-level pass corresponds to the recorded retention decision: all required items are 4--5 and no must-fix issue. A revision-required decision groups the action-required cases, including revise, eliminate, or escalate in artifacts with richer labels.

Extended screening checks grounding fidelity, scaffolding compatibility and foundation credibility, long-form potential, CS construct validity, openness/low constraint, and restraint of an optional ThinPlan. Grounded information must remain preserved, while added narrative material must remain compatible with it. The starting conditions must support development without prescribing the plot or a particular CS realization; language resources must permit meaningful adaptation.

\paragraph{Annotator panel and protocol.}
The same three annotators carried out all human annotation and rating work
reported in this appendix and in Appendix~\ref{app:human}:
Chinese--English bilingual postgraduate students, all native Chinese speakers
with high English proficiency who had passed CET-6, and all with prior
experience in linguistic annotation and in literary analysis. For prompt
screening, two of them reviewed each candidate independently and the third
adjudicated splits across the 1--3/4--5 boundary; a 4-versus-5 difference alone
did not. For output rating, all three scored the same sampled narratives
against the rubric, and the concordance reported in Appendix~\ref{app:human}
is computed over those three sets of scores.

\subsection{Recorded Screening Counts}
Machine checks cover structural completeness, Seed binding, visible/internal-field separation, fixed task constraints, source-copy risk, and near-duplicate prompts, with character 3-gram Jaccard similarity at least .85 flagging redundancy for review. All retained prompts satisfy the Jaccard-based redundancy criterion. Screening and revising the candidate prompts yields the final sets: 407 Focused prompts and 400 Extended prompts.

\subsection{Agreement}
The same three annotators described above conducted the prompt-screening evaluation.

Candidate construction yields 855 prompts, 415 Focused and 440 Extended. Against those denominators, binary human-screening agreement is Gwet's AC1 = .975 for Focused
(415 matched pairs; unrounded estimate .9750, 95\% CI [.9573, .9901])
and .838 for Extended (440 matched pairs; .8378, 95\% CI [.7853,
.8836]). The retained intervals use paired candidate-level percentile
bootstrap with 10,000 replicates. Prompts requiring revision were subsequently corrected and re-screened before final retention. Screening and revision then reduce these candidate pools to the final sets of 407 Focused and 400 Extended prompts.

%% file: appendix_support/results.tex
\section{Main Model-Level Results}
\label{app:results}

\subsection{Set Coverage and Scoring}
Table~\ref{tab:overall-performance} is produced by re-running the aggregation rules of Appendix~\ref{app:rubric} over the recorded event judgments.

\noindent\begin{minipage}{\columnwidth}
\centering
\input{appendix_support/generated_uncertainty/family_aggregate.tex}
\par\vspace{4pt}
\input{appendix_support/generated/focused_whole.tex}
\end{minipage}

The scored corpus contains 4,884 Focused articles and 4,800 Extended articles. All twelve models are scored in both settings, and every included event has valid binary D1 decisions from both judges.

\begin{figure*}[t]
\centering
\includegraphics[width=\textwidth]{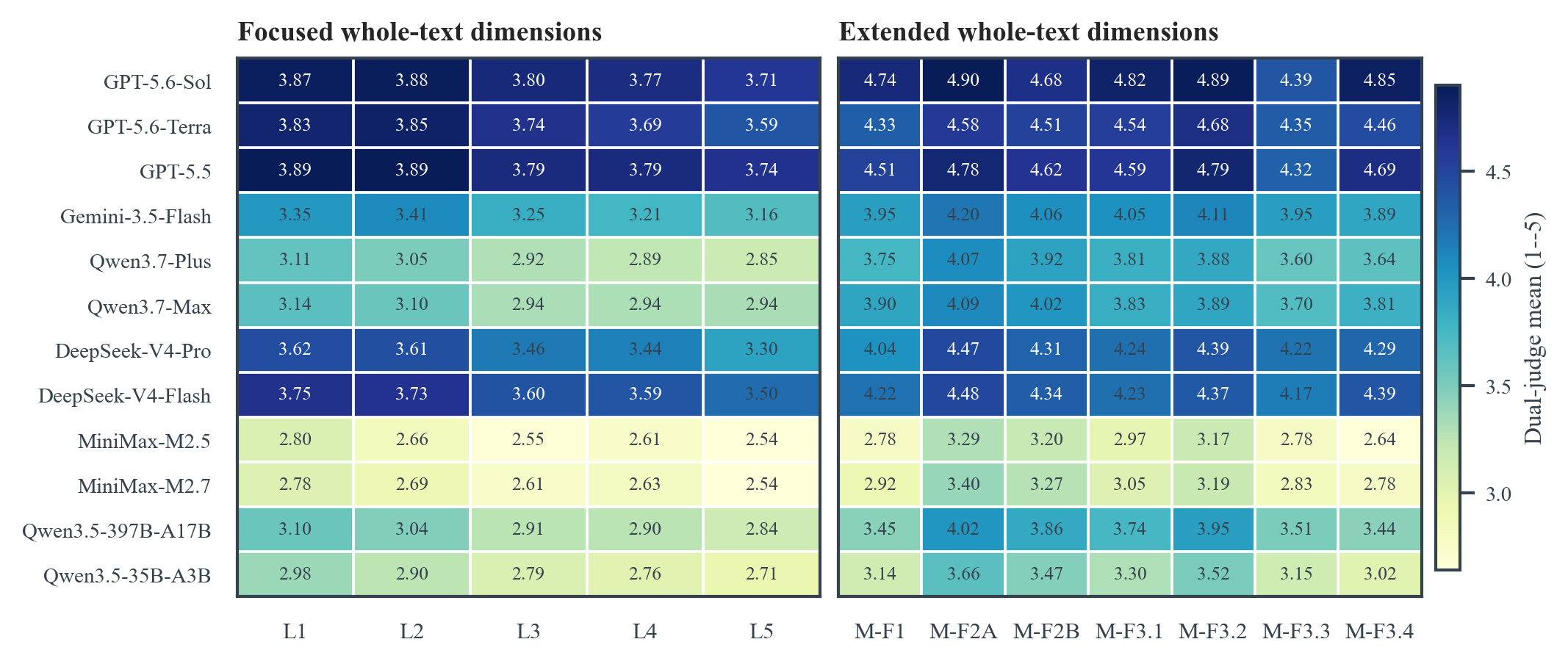}
\caption{Per-dimension model profiles for the two settings. The rightmost extended dimension group separates multilingual-integration dimensions from general narrative dimensions. Values are dual-judge means on the 1--5 dimension scale.}
\label{fig:result-dimension-profile}
\end{figure*}

Within each task, D1, D2, D3, and Whole are averaged over the articles, and their sum is the within-setting composite $\Sigma$ (/20). Table~\ref{tab:focused-whole} and Table~\ref{tab:extended-whole} report the whole-text components behind the dimension means of the main table, and Figure~\ref{fig:result-dimension-profile} shows the same judgments as a per-dimension profile. The profile makes the task-specific difference in separation visible: models that are close on general narrative execution can still differ substantially on the multilingual-integration dimensions.

Table~\ref{tab:family-aggregate} provides descriptive aggregates of the evaluated model instances grouped by model family.

\begin{figure*}[t]
\centering
\includegraphics[width=\textwidth]{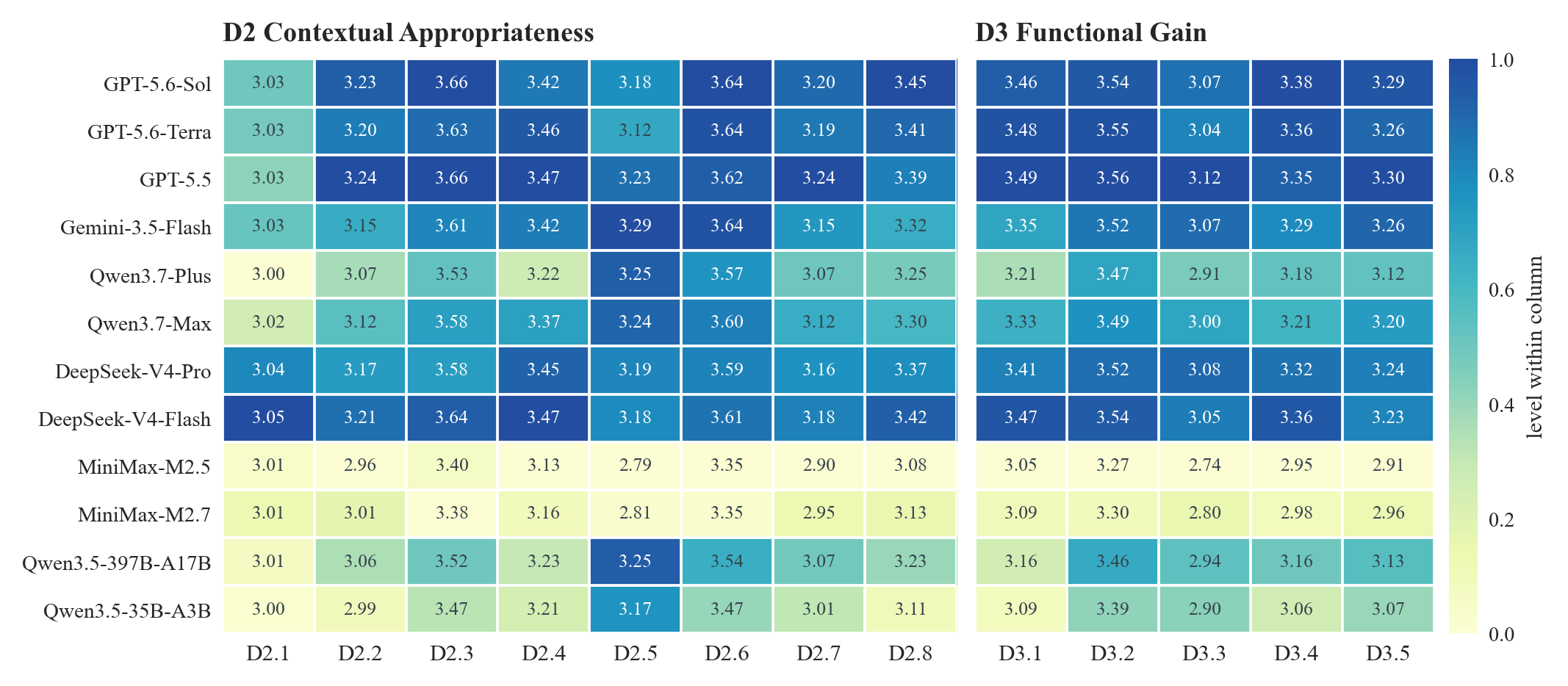}
\caption{Focused Narrative subdimension profiles for the twelve models. Columns are the eight D2 and five D3 subdimensions defined in Appendix~\ref{app:rubric}; cell values are dual-judge means on the original 1--5 scale, and each column is shaded against its own range so that darker cells mark the higher-scoring models within that subdimension.}
\label{fig:focused-subdimensions}
\end{figure*}

\input{appendix_support/generated/extended_whole.tex}

%% file: appendix_support/generated_uncertainty/family_aggregate.tex
\centering\scriptsize
\setlength{\tabcolsep}{3pt}

\begin{tabular}{lrrr}
\toprule
Family & Models & Focused /20 & Extended /20 \\
\midrule
GPT & 3 & 15.483 & 17.489 \\
DeepSeek & 2 & 15.208 & 16.765 \\
Gemini & 1 & 14.893 & 16.471 \\
Qwen & 4 & 14.341 & 15.619 \\
MiniMax & 2 & 13.671 & 13.637 \\
\bottomrule
\end{tabular}
\captionof{table}{Descriptive aggregates of the evaluated model instances grouped by model family. Each group mean is the unweighted mean of the evaluated model instances shown for that family. Group sizes differ, and these descriptive aggregates should not be interpreted as estimates of family-level population performance.}
\label{tab:family-aggregate}

%% file: appendix_support/generated/focused_whole.tex
\centering\scriptsize
\setlength{\tabcolsep}{3pt}

\begin{tabular}{lrrrrr}
\toprule
Model & L1 & L2 & L3 & L4 & L5 \\
\midrule
GPT-5.6-Sol & 3.869 & 3.881 & 3.797 & 3.773 & 3.711 \\
GPT-5.6-Terra & 3.829 & 3.849 & 3.743 & 3.690 & 3.588 \\
GPT-5.5 & 3.894 & 3.892 & 3.786 & 3.785 & 3.736 \\
Gemini-3.5-Flash & 3.351 & 3.405 & 3.251 & 3.209 & 3.163 \\
Qwen3.7-Plus & 3.114 & 3.053 & 2.924 & 2.891 & 2.851 \\
Qwen3.7-Max & 3.144 & 3.102 & 2.940 & 2.939 & 2.945 \\
DeepSeek-V4-Pro & 3.623 & 3.608 & 3.462 & 3.442 & 3.301 \\
DeepSeek-V4-Flash & 3.746 & 3.735 & 3.598 & 3.586 & 3.502 \\
MiniMax-M2.5 & 2.795 & 2.657 & 2.548 & 2.609 & 2.536 \\
MiniMax-M2.7 & 2.784 & 2.694 & 2.609 & 2.631 & 2.537 \\
Qwen3.5-397B-A17B & 3.101 & 3.038 & 2.905 & 2.903 & 2.844 \\
Qwen3.5-35B-A3B & 2.978 & 2.900 & 2.787 & 2.756 & 2.711 \\
\bottomrule
\end{tabular}
\captionof{table}{Focused whole-text dimension means (/5). IDs and definitions are in Appendix~\ref{app:rubric}.}
\label{tab:focused-whole}

%% file: appendix_support/generated/extended_whole.tex
\begin{table*}[t]
\centering\scriptsize
\setlength{\tabcolsep}{3pt}

\begin{tabular}{lrrrrrrr}
\toprule
Model & M-F1 & M-F2A & M-F2B & M-F3.1 & M-F3.2 & M-F3.3 & M-F3.4 \\
\midrule
GPT-5.6-Sol & 4.737 & 4.902 & 4.679 & 4.819 & 4.886 & 4.391 & 4.851 \\
GPT-5.6-Terra & 4.330 & 4.584 & 4.514 & 4.544 & 4.684 & 4.348 & 4.455 \\
GPT-5.5 & 4.510 & 4.777 & 4.621 & 4.595 & 4.786 & 4.315 & 4.691 \\
Gemini-3.5-Flash & 3.953 & 4.201 & 4.058 & 4.055 & 4.110 & 3.954 & 3.893 \\
Qwen3.7-Plus & 3.751 & 4.074 & 3.915 & 3.809 & 3.876 & 3.595 & 3.636 \\
Qwen3.7-Max & 3.896 & 4.091 & 4.018 & 3.834 & 3.894 & 3.697 & 3.808 \\
DeepSeek-V4-Pro & 4.040 & 4.471 & 4.306 & 4.237 & 4.388 & 4.216 & 4.293 \\
DeepSeek-V4-Flash & 4.224 & 4.482 & 4.339 & 4.232 & 4.369 & 4.168 & 4.388 \\
MiniMax-M2.5 & 2.781 & 3.289 & 3.204 & 2.970 & 3.174 & 2.779 & 2.643 \\
MiniMax-M2.7 & 2.922 & 3.401 & 3.271 & 3.046 & 3.194 & 2.831 & 2.781 \\
Qwen3.5-397B-A17B & 3.449 & 4.019 & 3.861 & 3.741 & 3.950 & 3.506 & 3.445 \\
Qwen3.5-35B-A3B & 3.138 & 3.656 & 3.471 & 3.303 & 3.515 & 3.151 & 3.021 \\
\bottomrule
\end{tabular}
\caption{Extended whole-text dimension means (/5). IDs and definitions are in Appendix~\ref{app:rubric}.}
\label{tab:extended-whole}
\end{table*}

%% file: appendix_support/analyses.tex
\section{Supplementary Evaluation Results and Additional Analyses}
\label{app:analysis}

\subsection{Judge Ablation and Ranking Robustness}
\label{app:judge_ablation}
Table~\ref{tab:judge_ablation} ranks the same twelve models from each primary-judge
score set after the current article and task-total aggregation. Dual is the article-level
judge mean. Spearman's $\rho$ correlates model ranks; absolute shifts compare matching
models, using average ranks for ties.

\input{appendix_support/generated/judge_ablation.tex}

Focused GPT-side rankings equal Dual; both other comparisons have $\rho=.993$,
mean shift .167, and maximum shift one. Extended DeepSeek-side rankings equal
Dual; both GPT-side comparisons have $\rho=.986$, mean shift .333, and maximum
shift one.

\begin{figure*}[tb]
\centering
\includegraphics[width=\textwidth]{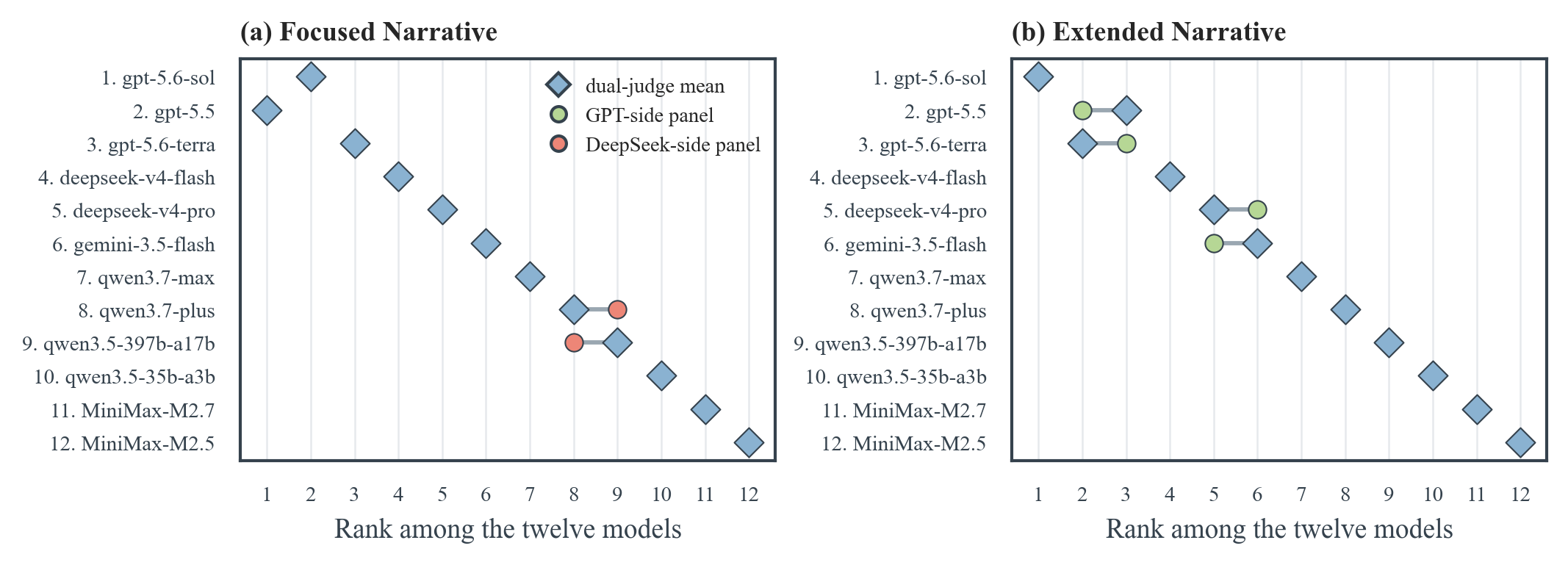}
\caption{Judge-side ranking shift for the twelve models under the benchmark
aggregation rules. The markers give the rank assigned by the GPT-side judge,
the DeepSeek-side judge, and their article-level mean.}
\label{fig:judge-ablation-shift}
\end{figure*}

\subsection{Constructor and Annotation Sensitivity}
\label{app:pipeline-sensitivity}

We examine whether the benchmark results are sensitive to the models used in two
upstream stages: Prompt construction and event annotation. We cross GPT- and
Qwen-based Prompt constructors with GPT- and Qwen-based event annotators in a
$2\times2$ design. Both stages instantiate the same model configurations as the
generation and judging stages, queried with the same decoding and reasoning
settings: GPT-5.6-Sol for the GPT-based arms and Qwen3.7-Max for the Qwen-based
arms (temperature $0.7$, top-$p$ $0.7$). For each configuration, we recompute
model performance using the same D1+D2+D3+Whole composite (/20) as in the
main analysis and compare the resulting ranking of the twelve generation models
with the original ranking.

The sensitivity panel is an independent rerun of the main experiment and covers
twenty Seeds. The two Prompt-constructor arms use separately reconstructed prompts
and regenerated articles, but within each constructor arm the GPT- and Qwen-based
event annotators are applied to the same generated narratives. The configuration
matching the main experiment therefore does not reproduce the main ranking
exactly; agreement below unity also reflects the smaller Seed sample and the
regeneration of prompts and articles.

\begin{table}[t]
\centering
\small
\setlength{\tabcolsep}{3pt}

\begin{tabular}{lcc}
\toprule
\textbf{Prompt constructor} &
\textbf{GPT annotator} &
\textbf{Qwen annotator} \\
\midrule
GPT &
\begin{tabular}[c]{@{}c@{}}F .951\\E .965\end{tabular} &
\begin{tabular}[c]{@{}c@{}}F .888\\E .979\end{tabular} \\
\midrule
Qwen &
\begin{tabular}[c]{@{}c@{}}F .958\\E .916\end{tabular} &
\begin{tabular}[c]{@{}c@{}}F .937\\E .972\end{tabular} \\
\bottomrule
\end{tabular}
\caption{Rank agreement with the original main-experiment ranking under the
$2\times2$ Prompt-constructor $\times$ event-annotator sensitivity design. Each
cell reports Spearman's $\rho$ across the twelve generation models for Focused
(F) and Extended (E) Narrative.}
\label{tab:pipeline-cells}
\end{table}

Rank agreement remains high across all four configurations in both settings,
with Spearman's $\rho$ ranging from $.888$ to $.979$. The effect of annotator choice on rank agreement varies across Prompt
constructors and narrative settings.

The score-level analysis provides a complementary view. For Focused Narrative,
the estimated effects of constructor choice, annotator choice, and their
interaction are small, with all bootstrap confidence intervals spanning zero.
For Extended Narrative, changing the event annotator corresponds to a pooled
mean score difference of $+0.745$ points on the /20 scale across the
article-matched within-constructor contrasts (95\% CI $[0.528,0.931]$). The
pipeline-level constructor contrast is $-0.601$ (95\% CI $[-1.552,0.377]$),
and the constructor--annotator interaction is $+0.636$
(95\% CI $[-0.026,1.262]$).

Across the four configurations, upstream model choice can affect absolute
scores while model rankings remain broadly stable. The twenty-Seed panel
provides a targeted assessment of upstream model dependence.

\subsection{Licensing Evidence Intervention}
\label{app:rescue}
We conduct a targeted intervention on 36 initially failed Licensing checks,
covering all twelve generation models, three articles per model, and equal
numbers of participant/role and world checks. For each target, we add one
authoritative fact to the Prompt that directly supplies the participant- or
world-level information required to license the corresponding event. The
generated narrative, fixed event manifest, and all other Prompt information
remain unchanged. All 36 checks subsequently change from FAIL to PASS.

This intervention directly manipulates the evidence relevant to Licensing
while holding the judged event and its realization fixed. The uniform
FAIL-to-PASS transition shows that Licensing judgments respond in the
expected direction when the task context is changed from lacking the
relevant licensing information to explicitly providing it. This targeted
analysis therefore provides a construct-sensitivity check for the operational
Licensing criterion.

\subsection{Human Validation}
\label{app:human}
The same three annotators each evaluated 180 Focused and 144 Extended outputs.
Focused results are Kendall's $W=.718$, ICC(A,1)=.864, and Spearman's $\rho=.863$;
Extended results are $W=.810$, ICC(A,1)=.816, and $\rho=.804$. Kendall's $W$ measures
three-rater concordance; ICC(A,1) and $\rho$ compare their mean with the dual-judge
article mean.

\input{appendix_support/generated_uncertainty/cross_judge.tex}

We also validate the event-discovery stage with two independent bilingual annotators. The first analysis examines candidate inclusion: whether each highlighted English candidate constitutes genuine code-switching. English material is labeled CS when it forms part of the linguistic realization of the Chinese--English discourse itself, and non-CS when it instead consists of a lexicalized borrowing, proper name, quotation or linguistic mention, printed or screen-displayed text, decorative material, an incorrectly extracted fragment, or other multilingual material outside the code-switching discourse.

For readability, adjacent candidates can be shown as continuous runs, while the individual candidate remains the unit of judgment; run-level decisions are mapped back to their member candidates with candidate-level corrections where necessary. Both annotators are blinded to GPT-5.6-Sol's classifications and to each other's judgments. We compare GPT-5.6-Sol separately with each annotator. Because the 108-narrative sample is stratified by the number of GPT-classified non-CS candidates, candidate-level F1 is post-stratified using candidate-mass weights from the full evaluation frame.

We next examine event grouping, since downstream evaluation operates on code-switching events rather than isolated candidates. On a targeted 48-narrative subset, the same two annotators independently group their CS candidates into events. Candidates belong to the same event when they form one continuous occurrence of code-switching; a return to Chinese, a change of speaker turn, or the onset of a distinct switching occurrence marks a new event.

For each GPT--annotator comparison, both event partitions are restricted to candidates classified as CS by both sides. Agreement is then computed over this shared candidate set. Grouping F1 measures whether pairs of candidates are assigned to the same event; Boundary F1 compares the event boundaries induced by the two groupings; and Exact reports the proportion of articles with identical projected partitions.

\begin{table}[t]
\centering
\small
\setlength{\tabcolsep}{0.8pt}

\begin{tabular}{llccc}
\toprule
\textbf{Setting} & \textbf{Ref.} & \textbf{Grouping F1 [95\% CI]} & \textbf{Boundary F1} & \textbf{Exact} \\
\midrule
Focused  & A1 & .919 [.820, 1.000] & .980 & .875 \\
Focused  & A2 & .968 [.917, 1.000] & .990 & .917 \\
Extended & A1 & .998 [.994, 1.000] & .988 & .958 \\
Extended & A2 & .999 [.997, 1.000] & .990 & .917 \\
\bottomrule
\end{tabular}
\caption{Agreement between GPT-5.6-Sol's recorded event grouping and the two human annotators over candidates identified as CS by both sides. Grouping F1 compares same-event membership; Boundary F1 compares grouping-induced event boundaries; Exact is the proportion of articles with identical projected event partitions.}
\label{tab:event-grouping-audit}
\end{table}

\subsection{Extended Evidence Fidelity}
\label{app:evidence-fidelity}

Extended Narrative uses evidence records constructed before quality judging to summarize character, relationship, narrative-world, and narrative-state information. We audit these records on 36 narratives, sampling three outputs from each of the twelve generation models from the evaluation frame excluding the human-rating subset. The records contain 8,209 atomic assertions. Two independent human auditors assess each assertion for factual support in the narrative and, where applicable, whether it is assigned to the correct point in the narrative progression. Disagreements are resolved by a third human auditor, and confidence intervals use 10,000 narrative-cluster bootstrap resamples.

Across the 8,209 assertions covered by both auditor arms, adjudicated factual accuracy is 98.40\% [97.86, 98.89]. Temporal accuracy is 99.26\% [98.99, 99.52] over the 6,062 assertions for which a temporal judgment is applicable. Accuracy remains high across all five annotation classes (Table~\ref{tab:evidence-fidelity}).

\begin{table}[t]
\centering
\small
\setlength{\tabcolsep}{4pt}

\begin{tabular}{lrr}
\toprule
\textbf{Annotation class} &
\textbf{Factual acc.} &
\textbf{Temporal acc.} \\
\midrule
Character profile      & 98.93\% & 96.98\% \\
Character state        & 97.77\% & 99.44\% \\
Narrative arc          & 99.50\% & 99.61\% \\
Relationship state     & 96.59\% & 99.24\% \\
World state            & 99.19\% & 99.18\% \\
\bottomrule
\end{tabular}
\caption{Adjudicated fidelity of the evidence records used in Extended Narrative. Temporal accuracy is computed over assertions for which temporal placement is applicable.}
\label{tab:evidence-fidelity}
\end{table}

The audited evidence records closely reflect both the narrative content and its temporal development.

\subsection{Evaluator--Generator Self- and Family Preference}
\label{app:self-preference}

Because GPT-5.6-Sol and DeepSeek-V4-Flash serve both as generation models and as
evaluators, we examine evaluator--generator overlap at both the checkpoint and
model-family levels. The analysis uses the same human-validation subsets as
Section~\ref{sec:rater-agreement}: 180 Focused Narrative outputs and 144 Extended
Narrative outputs. This analysis isolates evaluator overlap; upstream model
dependence is assessed separately through the constructor--annotator sensitivity
analysis, final-Prompt human screening, and the human audits of event discovery
and Extended Narrative evidence records.

For each article $i$ and evaluator $j$, we compute the evaluator residual
relative to the three-rater human consensus,
\[
r_{ij}
=
S_{ij}^{\mathrm{judge}}
-
S_i^{\mathrm{human}}.
\]
We first compare the mean residual on the evaluator's own generations with that on
all non-self generations. We then separate exact-self, same-family, and
other-family generations for the comparison that shows a positive signal.

\begin{table*}[t]
\centering
\small
\setlength{\tabcolsep}{5pt}

\begin{tabular}{llrrrrc}
\toprule
\textbf{Setting} & \textbf{Evaluator} &
\textbf{Self $n$} & \textbf{Non-self $n$} &
\textbf{Self residual} & \textbf{Non-self residual} &
\textbf{Self $-$ non-self [95\% CI]} \\
\midrule
Focused  & GPT-5.6-Sol
& 15 & 165
& \phantom{$-$}0.548
& \phantom{$-$}0.193
& \phantom{$-$}0.355 [0.100, 0.595] \\
Focused  & DeepSeek-V4-Flash
& 15 & 165
& $-$0.351
& $-$0.238
& $-$0.112 [$-$0.409, 0.248] \\
Extended & GPT-5.6-Sol
& 12 & 132
& $-$1.236
& $-$1.337
& \phantom{$-$}0.101 [$-$1.370, 1.482] \\
Extended & DeepSeek-V4-Flash
& 12 & 132
& \phantom{$-$}0.632
& \phantom{$-$}1.096
& $-$0.463 [$-$1.333, 0.524] \\
\bottomrule
\end{tabular}
\caption{Human-anchored evaluator--generator overlap diagnostic.
Residuals are evaluator article totals minus the three-rater human consensus.
The final contrast compares exact-self outputs with all non-self outputs.
Confidence intervals use 2,000 article-level bootstrap resamples.}
\label{tab:self-preference-overall}
\end{table*}

Table~\ref{tab:self-preference-overall} shows a positive interval excluding zero
only for GPT-5.6-Sol in Focused Narrative. We therefore decompose this signal
into \emph{exact self} (GPT-5.6-Sol), \emph{same family} (GPT-5.5 and
GPT-5.6-Terra), and \emph{other family} (the remaining generation models).

\begin{table}[t]
\centering
\small
\setlength{\tabcolsep}{4pt}

\begin{tabular}{lcc}
\toprule
\textbf{Contrast} & \textbf{Estimate} & \textbf{95\% CI} \\
\midrule
Exact self $-$ other family
    & \phantom{$-$}0.450 & [0.170,\;0.682] \\
Same family $-$ other family
    & \phantom{$-$}0.522 & [0.378,\;0.667] \\
Exact self $-$ same family
    & $-$0.072 & [$-$0.345,\;0.161] \\
\bottomrule
\end{tabular}
\caption{Decomposition of the Focused Narrative GPT evaluator signal.
Contrasts are differences in evaluator-minus-human article-level residuals.}
\label{tab:self-family}
\end{table}

The Focused GPT signal is attributable to family-level affinity. Exact-self and
same-family outputs both have higher residuals than other-family outputs, while
the exact-self and same-family groups are statistically indistinguishable
($-0.072$, 95\% CI $[-0.345,0.161]$).

We next test whether evaluator--generator overlap materially affects the
cross-model comparison. We construct a leave-self-judge-out score in which
GPT-5.6-Sol generations are evaluated using only DeepSeek-V4-Flash,
DeepSeek-V4-Flash generations are evaluated using only GPT-5.6-Sol, and all
other generation models retain their original dual-evaluator mean. Recomputing
the twelve-model rankings shows that the model ordering is unchanged in both
Focused and Extended Narrative relative to the original dual-evaluator ranking
(Spearman's $\rho=1.000$ in both settings).

We additionally exclude GPT-family judge scores for GPT-family generations,
while all other generation models retain their original dual-evaluator mean.
Under this family-excluded panel, the model ordering likewise remains unchanged
in both settings: GPT-5.5 remains the highest-scoring model in Focused Narrative,
and GPT-5.6-Sol remains the highest-scoring model in Extended Narrative. Thus,
although the human-anchored residual analysis identifies a Focused GPT
family-level affinity signal, neither evaluator-overlap robustness check changes
the cross-model ordering.

\input{appendix_support/generated/position_swap_page.tex}

\subsection{Position Swaps}
\label{app:position-swaps}

The analysis uses 36 Extended narratives, with one LOW-, MID-, and
HIGH-scoring eligible narrative sampled from each of the twelve generation
models. Eligibility requires at least 16 previously identified semantic CS events so that
eight disjoint event pairs can be formed.

Within each narrative, identified CS events are ordered by textual midpoint and
used to form eight disjoint long-distance pairs. These pairs are ranked by
normalized midpoint distance, and the 2-, 4-, 6-, and 8-pair conditions
cumulatively exchange the exact event spans in that order. The event-text
multiset and all non-event text remain unchanged.

As a control, we apply the same perturbation schedule to Chinese-only clause
pairs that do not overlap with identified CS events. The same cumulative 2-, 4-,
6-, and 8-pair structure is used.

Table~\ref{tab:position-swap} reports the 36-narrative analysis judged by
GPT-5.6-Sol and DeepSeek-V4-Flash. Baseline and all CS and Chinese conditions
share the same complete-case subset, and comparisons are paired within
narrative. Across 10{,}000 paired percentile-bootstrap resamples, the
Chinese-minus-CS difference at eight pairs is $+0.236$ [$0.071$, $0.411$]
whole-text, $+0.594$ [$0.375$, $0.823$] multilingual, and $-0.241$
[$-0.384$, $-0.093$] general.

\Needspace{14\baselineskip}
\subsection{Aggregation Sensitivity and Within-Setting Model Comparisons}
\label{app:aggregation-sensitivity}

We test the sensitivity of the within-setting $\Sigma$ model ordering to three
alternative aggregations: removing D1, assigning D1 half weight, and replacing
the score composite with the average of the four dimension-specific model ranks.

Removing D1 or assigning it half weight leaves the model ordering unchanged in
both narrative settings. Rank-averaging remains perfectly rank-correlated with
the Focused baseline and highly consistent with the Extended baseline
($\rho=.977$), with differences confined to closely spaced models
(Table~\ref{tab:aggregation-sensitivity}). The highest-ranked model also remains
unchanged under all three alternatives: GPT-5.5 in Focused Narrative and
GPT-5.6-Sol in Extended Narrative.

\begin{table*}[t]
\centering
\small
\setlength{\tabcolsep}{5pt}

\begin{tabular}{llrrl}
\toprule
\textbf{Setting} & \textbf{Aggregation} & $\boldsymbol{\rho}$ &
\textbf{Max $|\Delta r|$} & \textbf{Top model} \\
\midrule
Focused & Drop D1 & 1.000 & 0 & GPT-5.5 \\
Focused & D1 half weight & 1.000 & 0 & GPT-5.5 \\
Focused & Rank average & 1.000 & 0.5 & GPT-5.5 \\
Extended & Drop D1 & 1.000 & 0 & GPT-5.6-Sol \\
Extended & D1 half weight & 1.000 & 0 & GPT-5.6-Sol \\
Extended & Rank average & .977 & 1.5 & GPT-5.6-Sol \\
\bottomrule
\end{tabular}
\caption{Sensitivity of within-setting model ordering to alternative aggregation
schemes. Spearman's $\rho$ is computed against the equal-weight $\Sigma$
ordering. Average ranks are used for ties. For the rank-average rows, Max
$|\Delta r|$ denotes the change in the underlying average-rank value rather
than displacement in final leaderboard position.}
\label{tab:aggregation-sensitivity}
\end{table*}

\paragraph{Licensing spread.}
Licensing shows substantially less between-model variation in Focused Narrative
than the other reported dimensions. Across the twelve models, the Focused range
is .061 for D1, compared with .282 for D2, .380 for D3, and 1.190 for Whole
(Table~\ref{tab:dimension-spread}). The corresponding D1 standard deviation
across models is .021. Extended Narrative shows greater D1 variation, with a
range of .487, while the D2, D3, and Whole ranges are .942, 1.120, and 1.775.
Despite the near-ceiling behavior of Focused D1, removing D1 leaves the Focused
model ordering unchanged; the Extended ordering is likewise unchanged when D1
is removed.

\begin{table}[t]
\centering
\small
\setlength{\tabcolsep}{5pt}

\begin{tabular}{lrrrr}
\toprule
\textbf{Setting} & \textbf{D1} & \textbf{D2} & \textbf{D3} & \textbf{Whole} \\
\midrule
Focused & .061 & .282 & .380 & 1.190 \\
Extended & .487 & .942 & 1.120 & 1.775 \\
\bottomrule
\end{tabular}
\caption{Across-model range of the reported dimensions in each narrative
setting. Each value is the maximum model mean minus the minimum model mean.}
\label{tab:dimension-spread}
\end{table}

\paragraph{Within-setting pairwise uncertainty.}
We additionally compute paired Seed-clustered bootstrap differences between all
66 model pairs separately within each setting. The analysis uses the same
clustered resampling scheme as the model-level intervals: 546 Seed clusters,
10{,}000 replicates, and one shared multiplicity vector within each replicate.
All comparisons use the dual-judge article means underlying the reported
within-setting $\Sigma$ scores. Holm correction is applied across the 66
comparisons within each setting.

In Focused Narrative, 65 of 66 percentile-bootstrap intervals exclude zero,
and 63 of 66 comparisons remain significant after Holm correction. In Extended
Narrative, 64 of 66 intervals exclude zero, and all 64 remain significant after
correction. Thus most model pairs are clearly separated within each setting,
while a small number of closely scoring pairs do not support a stronger
pairwise ordering after multiplicity correction. Table~\ref{tab:pairwise-close}
lists the five comparisons that are not significant after Holm correction.

For GPT-5.6-Sol versus GPT-5.5 and MiniMax-M2.5 versus MiniMax-M2.7 in Focused
Narrative, the unadjusted bootstrap intervals already exclude zero, but the
comparisons fall just above the $.05$ threshold after correcting across all
66 pairs.

\subsection{Source and Seed Dependency Robustness}
\label{app:source-dependency}

We examine whether within-setting model comparisons are sensitive to source
provenance and cross-setting Seed reuse. For each analysis, model scores are
recomputed from the original dual-judge article scores using the same within-setting
$\Sigma=D1+D2+D3+\mathrm{Whole}$ aggregation as in the main experiment. Rank agreement
is measured with Spearman's $\rho$ across the twelve generation models.
Confidence intervals use 2,000 prompt-clustered bootstrap resamples.

We compare ASCEND-derived versus SwitchLingua-derived articles and Seeds shared
by the two narrative settings versus task-specific Seeds. Table~\ref{tab:source-dependency}
reports the resulting rank agreement.

The two source corpora are unevenly represented in the prompt panels: ASCEND
supplies 76 of the 407 Focused prompts (18.7\%) and 70 of the 400 Extended
prompts (17.5\%), with SwitchLingua supplying the remaining 331 and 330.
Differences in absolute scores across the two source corpora are what the
comparisons below measure, and the source-balanced reweighting at the end of
this section reports the ordering when the two corpora contribute equally.

Rank agreement remains high across both comparisons
($\rho=.986$--$.993$). Restricting the evaluation by source corpus or by
shared versus task-specific Seeds changes the ordering only through adjacent
rank exchanges, with a maximum shift of one position.

Source provenance shows somewhat greater score-level variation than Seed reuse.
Across the 24 model-by-setting ASCEND-minus-SwitchLingua contrasts, 9 have 95\%
bootstrap intervals excluding zero, with gaps ranging from $-0.139$ to $+0.417$
points on the /20 scale. By contrast, none of the 24 model-level contrasts for
shared versus task-specific Seeds has an interval excluding zero. Thus the source
corpus can shift absolute score levels for some models, while the relative
ordering remains highly stable across the tested source and Seed compositions.

Finally, to check whether the unequal source proportions drive the within-setting
ordering, we recompute each model's composite with ASCEND and SwitchLingua weighted
equally,
$\Sigma_{\mathrm{bal}}=\tfrac{1}{2}\Sigma_{\mathrm{ASCEND}}+\tfrac{1}{2}\Sigma_{\mathrm{SWL}}$.
The reweighted composite reproduces the unweighted ranking in Extended Narrative
($\rho=1.000$, 95\% CI $[0.993, 1.000]$) and differs from it in Focused Narrative
only by the adjacent exchange of ranks 8 and 9 ($\rho=0.993$, 95\% CI
$[0.986, 1.000]$).

\subsection{Uncertainty and Ranking Robustness}
\label{app:uncertainty}
Tables~\ref{tab:association-ci}, \ref{tab:model-ci}, and \ref{tab:difficulty-ci}
report cross-judge associations, model-level confidence intervals, and
prompt-difficulty gaps, respectively. Across 10{,}000
replicates, we resample 546 Seeds while keeping each sampled Seed's prompts and
articles together; model means and ranks share draws across settings, whereas
association intervals resample within setting. Percentile limits use the 2.5th
and 97.5th percentiles.

Source-group resampling retains each group's prompts (Focused: 407/350
prompts/groups; Extended: 400/340); rank intervals and top-model/top-three
retention are unchanged, while sampling standard deviations rise 3.3--3.7\%
(Table~\ref{tab:source-group-bootstrap}).

To examine sensitivity to prompt difficulty, we use total scores
computed with unweighted local D2/D3 means from six representative models:
GPT-5.6-Sol, DeepSeek-V4-Flash, Gemini-3.5-Flash, MiniMax-M2.7,
Qwen3.7-Max, and Qwen3.5-35B-A3B. Prompts in the lowest- and
highest-scoring quartiles are designated hard and easy, respectively.
The remaining six models, GPT-5.6-Terra, GPT-5.5, DeepSeek-V4-Pro,
MiniMax-M2.5, Qwen3.7-Plus, and Qwen3.5-397B-A17B, are used
to compare dimension scores between these groups (Table~\ref{tab:difficulty-ci}).

Among the seven Extended Whole-text dimensions, M-F3.4 has the largest
cross-model separation (model-mean SD 0.742; range 2.209). On the raw
1--5 scale, M-F2B has the largest easy-to-hard decline (0.4333), followed
by M-F3.4 (0.3025). For each dimension, we divide the easy--hard gap by
the pooled within-group SD,
$\sqrt{(s_{\mathrm{hard}}^2+s_{\mathrm{easy}}^2)/2}$,
computed from prompt-level scores that average the six validation models.
M-F3.4 shows the largest standardized decline (1.336), followed by M-F2B
(1.216); the mean raw decline across M-F3.1--M-F3.3 is 0.088.

Repeating the hard/easy split seven times, each omitting one Whole-text
dimension, keeps every gap positive. For M-F3.4, the gap changes from 0.3025 to
0.3017 (99/100 hard prompts retained), and the M-F1/M-F3.4 versus
M-F3.1--M-F3.3 contrast changes from 0.213 to 0.210.

For the within-article association in Section~\ref{sec:salience-quality}, we
globally rank each variable within judge, then subtract its article mean.
We correlate judge $j$'s centered event salience $s_{aej}$ with judge $j'$'s
centered dimension-$d$ score $D^{(d)}_{aej'}$ in both directions:
\begin{equation}
\begin{split}
\tilde{x}_{aej}
&=x_{aej}-\frac{1}{|N_a|}\sum_{e'\in N_a}x_{ae'j},\\
r^{(d)}_{j\rightarrow j'}
&=\operatorname{Pearson}\!\left(\tilde{s}_{aej},\
\widetilde{D^{(d)}}_{aej'}\right),\qquad j\neq j',\\
r^{(d)}
&=\frac{1}{2}\left(r^{(d)}_{1\rightarrow2}+r^{(d)}_{2\rightarrow1}\right).
\end{split}
\end{equation}
Here $x$ denotes either ranked variable, $N_a$ is article $a$'s event set,
and $j'\neq j$. Pearson's $r$ pools the centered event pairs; the two reciprocal
correlations are averaged. Centering removes differences between articles
within a task, unlike Section~\ref{sec:local-to-whole}'s generation-model and
prompt-card fixed effects.

\subsection{Switching Behavior and Coverage}
\label{app:switching-coverage}

We characterize switching independently of quality. Table~\ref{tab:switching-coverage}
reports events per article, CS density per 1,000 non-whitespace code points,
and the proportion of article characters covered by identified CS-event spans.

Table~\ref{tab:switching-score-assoc} reports pooled and within-model
associations between switching propensity and benchmark scores.

Event count and CS density show weak associations with $\Sigma$ (event count:
$r=.032$ Extended, $-.052$ Focused; density: $r=-.123$ and $-.159$).
This pattern highlights the distinction between switching frequency and the
event-level qualities assessed by D1--D3: licensing, contextual fit, and
functional contribution.

\clearpage

\begin{table*}[t]
\centering\small
\input{appendix_support/generated_uncertainty/associations_body.tex}
\end{table*}

\input{appendix_support/generated_uncertainty/model_ci.tex}

\clearpage

\begin{table*}[t]
\centering
\small
\setlength{\tabcolsep}{4.5pt}

\begin{tabular}{lrrrrr}
\toprule
Model & Events/article & Median & CS density /1k & CS span (\%) & $\Sigma$ (/20) \\
\midrule
\multicolumn{6}{l}{\textit{Extended Narrative}} \\
\cmidrule(lr){1-6}
GPT-5.6-Sol          & 20.36 & 20 & 4.78 & 9.1  & 17.767 \\
GPT-5.6-Terra        & 11.39 & 11 & 2.94 & 5.5  & 17.352 \\
GPT-5.5              & 20.52 & 20 & 4.51 & 6.6  & 17.348 \\
DeepSeek-V4-Flash    & 17.32 & 17 & 4.54 & 11.2 & 16.962 \\
DeepSeek-V4-Pro      & 20.25 & 21 & 4.16 & 9.0  & 16.567 \\
Gemini-3.5-Flash     & 12.65 & 12 & 3.10 & 5.0  & 16.471 \\
Qwen3.7-Max          & 11.70 & 11 & 3.72 & 8.4  & 16.220 \\
Qwen3.7-Plus         & 11.00 & 10 & 3.42 & 7.6  & 16.065 \\
Qwen3.5-397B-A17B    & 15.44 & 14 & 3.33 & 3.6  & 15.491 \\
Qwen3.5-35B-A3B      & 13.54 & 12 & 3.06 & 4.5  & 14.698 \\
MiniMax-M2.7         & 18.67 & 18 & 5.76 & 16.0 & 13.818 \\
MiniMax-M2.5         & 15.31 & 14 & 4.31 & 11.0 & 13.456 \\
\midrule
\multicolumn{6}{l}{\textit{Focused Narrative}} \\
\cmidrule(lr){1-6}
GPT-5.5              & 6.99 & 7 & 8.09 & 12.4 & 15.537 \\
GPT-5.6-Sol          & 7.04 & 7 & 8.76 & 17.1 & 15.501 \\
GPT-5.6-Terra        & 6.19 & 6 & 9.70 & 17.9 & 15.410 \\
DeepSeek-V4-Flash    & 6.89 & 7 & 9.62 & 20.9 & 15.302 \\
DeepSeek-V4-Pro      & 8.35 & 8 & 12.21 & 24.3 & 15.114 \\
Gemini-3.5-Flash     & 5.53 & 5 & 7.74 & 15.5 & 14.893 \\
Qwen3.7-Max          & 5.84 & 6 & 8.76 & 18.1 & 14.540 \\
Qwen3.7-Plus         & 5.86 & 6 & 8.55 & 16.8 & 14.376 \\
Qwen3.5-397B-A17B    & 7.18 & 7 & 10.35 & 14.6 & 14.357 \\
Qwen3.5-35B-A3B      & 6.96 & 7 & 9.46 & 12.2 & 14.090 \\
MiniMax-M2.7         & 7.46 & 7 & 10.56 & 23.8 & 13.714 \\
MiniMax-M2.5         & 7.15 & 7 & 10.07 & 21.3 & 13.628 \\
\bottomrule
\end{tabular}
\caption{Switching behavior by generation model. CS density is the number of
discovered CS events per 1,000 non-whitespace code points; CS span is the mean
article-level proportion covered by identified CS-event spans.}
\label{tab:switching-coverage}
\end{table*}

\begin{table*}[t]
\centering
\input{appendix_support/generated_uncertainty/difficulty_body.tex}
\end{table*}

\clearpage

\begin{table*}[t]
\centering
\small
\setlength{\tabcolsep}{4pt}

\begin{tabular}{llrrrr}
\toprule
Measure & Basis & D2 & D3 & Whole & $\Sigma$ \\
\midrule
\multicolumn{6}{l}{\textit{Extended Narrative}} \\
Events  & pooled & -.026 & +.014 & +.163 & +.032 \\
Events  & within & -.105 & -.123 & +.030 & -.076 \\
Density & pooled & -.149 & -.145 & -.063 & -.123 \\
Density & within & -.049 & -.077 & +.026 & -.027 \\
\midrule
\multicolumn{6}{l}{\textit{Focused Narrative}} \\
Events  & pooled & -.141 & -.099 & -.002 & -.052 \\
Events  & within & -.082 & -.056 & -.010 & -.040 \\
Density & pooled & -.197 & -.160 & -.132 & -.159 \\
Density & within & -.110 & -.086 & -.076 & -.094 \\
\bottomrule
\end{tabular}
\caption{Article-level associations between switching propensity and benchmark
scores. Within-model correlations remove generation-model means before
computing the correlation.}
\label{tab:switching-score-assoc}
\end{table*}

\begin{table*}[t]
\centering
\small
\setlength{\tabcolsep}{5pt}

\begin{tabular}{llrrr}
\toprule
\textbf{Setting} & \textbf{Model pair} & $\boldsymbol{\Delta\Sigma}$ &
\textbf{95\% CI} & \textbf{Holm $p$} \\
\midrule
Focused & GPT-5.6-Sol -- GPT-5.5
& $-0.0355$ & $[-0.0670,-0.0040]$ & .0536 \\
Focused & Qwen3.7-Plus -- Qwen3.5-397B-A17B
& $+0.0193$ & $[-0.0343,0.0717]$ & .4648 \\
Focused & MiniMax-M2.5 -- MiniMax-M2.7
& $-0.0865$ & $[-0.1578,-0.0143]$ & .0504 \\
Extended & GPT-5.6-Terra -- GPT-5.5
& $+0.0046$ & $[-0.0899,0.0975]$ & .9064 \\
Extended & Gemini-3.5-Flash -- DeepSeek-V4-Pro
& $-0.0967$ & $[-0.2176,0.0257]$ & .2368 \\
\bottomrule
\end{tabular}
\caption{Within-setting model pairs that are not significant after Holm
correction over the 66 pairwise comparisons. Differences are row-first minus
row-second in the reported $\Sigma$ (/20).}
\label{tab:pairwise-close}
\end{table*}

\begin{table*}[t]
\centering
\small
\setlength{\tabcolsep}{4.5pt}

\begin{tabular}{llrrrr}
\toprule
\textbf{Setting} &
\textbf{Comparison} &
\textbf{Articles A / B} &
$\boldsymbol{\rho}$ &
\textbf{95\% CI} &
\textbf{Max $|\Delta r|$} \\
\midrule
Focused
& ASCEND vs.\ SwitchLingua
& 912 / 3,972
& .993
& [.979, 1.000]
& 1 \\

Focused
& Shared vs.\ task-specific Seeds
& 3,132 / 1,752
& .993
& [.979, 1.000]
& 1 \\
\midrule
Extended
& ASCEND vs.\ SwitchLingua
& 840 / 3,960
& .986
& [.944, 1.000]
& 1 \\

Extended
& Shared vs.\ task-specific Seeds
& 3,132 / 1,668
& .986
& [.979, 1.000]
& 1 \\
\bottomrule
\end{tabular}
\caption{Within-setting robustness of model ordering to source and Seed
composition. Spearman's $\rho$ compares the twelve-model rankings under the two
conditions in each row. Articles A and B are the scored articles belonging to
each side of the comparison, twelve per prompt; the prompt-level counts behind
them are reported in the text. Confidence intervals use prompt-clustered
bootstrap resampling.}
\label{tab:source-dependency}
\end{table*}

\begin{table*}[t]
\centering\small
\setlength{\tabcolsep}{6pt}

\begin{tabular}{lrrrr}
\toprule
Setting & Clusters & Widest rank & $P(\text{rank-1}$ & $P(\text{top-3}$ \\
 & (Seed/group) & interval & $\text{holds})$ & $\text{set holds})$ \\
\midrule
Focused  & 407 / 350 & 1 / 1 & .986 / .987 & 1.000 / 1.000 \\
Extended & 400 / 340 & 1 / 1 & 1.000 / 1.000 & 1.000 / 1.000 \\
\bottomrule
\end{tabular}
\caption{Source-group clustered bootstrap. Every quantity is reported for Seed
clustering (current intervals) and, after the slash, for source-group clustering.
Both settings are drawn jointly. The widest rank interval is the largest
$97.5\%$-minus-$2.5\%$ rank range across the twelve models.}
\label{tab:source-group-bootstrap}
\end{table*}

\clearpage

%% file: appendix_support/generated/judge_ablation.tex
\begin{table}[t]
\centering\scriptsize
\setlength{\tabcolsep}{3pt}

\begin{tabular}{lrrr}
\toprule
Task / comparison & $\rho$ & Mean $|\Delta r|$ & Max $|\Delta r|$ \\
\midrule
Focused: GPT vs. Dual & 1.000 & 0.000 & 0 \\
Focused: DeepSeek vs. Dual & 0.993 & 0.167 & 1 \\
Focused: GPT vs. DeepSeek & 0.993 & 0.167 & 1 \\
Extended: GPT vs. Dual & 0.986 & 0.333 & 1 \\
Extended: DeepSeek vs. Dual & 1.000 & 0.000 & 0 \\
Extended: GPT vs. DeepSeek & 0.986 & 0.333 & 1 \\
\bottomrule
\end{tabular}
\caption{Rank robustness on the same twelve models and current /20 task totals. Average ranks resolve ties.}
\label{tab:judge_ablation}
\end{table}

%% file: appendix_support/generated_uncertainty/cross_judge.tex
\begin{table}[tb]
\centering\scriptsize
\setlength{\tabcolsep}{3pt}

\begin{tabular}{lrrlrr}
\toprule
 & \multicolumn{2}{c}{D2} & & \multicolumn{2}{c}{D3} \\
\cmidrule(lr){2-3}\cmidrule(lr){5-6}
Quantity & Sol$\to$DS & DS$\to$Sol & Quantity & Sol$\to$DS & DS$\to$Sol \\
\midrule
Whole & 0.269 & 0.239 & Whole & 0.274 & 0.237 \\
M-F1 & 0.221 & 0.214 & M-F1 & 0.212 & 0.199 \\
M-F2A & 0.222 & 0.155 & M-F2A & 0.235 & 0.157 \\
M-F2B & 0.185 & 0.184 & M-F2B & 0.183 & 0.184 \\
M-F3.1 & 0.195 & 0.151 & M-F3.1 & 0.193 & 0.147 \\
M-F3.2 & 0.159 & 0.127 & M-F3.2 & 0.167 & 0.145 \\
M-F3.3 & 0.117 & 0.079 & M-F3.3 & 0.135 & 0.077 \\
M-F3.4 & 0.235 & 0.213 & M-F3.4 & 0.238 & 0.213 \\
\bottomrule
\end{tabular}
\caption{Cross-judge associations in Extended Narrative: one judge's event-level score against the other judge's whole-text score, in both directions. Sol$\to$DS reads from the event-level judge to the whole-text judge, and DS$\to$Sol reverses the two roles; the arrows mark the direction of the association, not a causal path. Section~\ref{sec:local-to-whole} reports their mean. Generation-model and prompt-card fixed effects are removed.}
\label{tab:cross-judge}
\end{table}

%% file: appendix_support/generated/position_swap_page.tex
\begin{figure*}[t]
\centering
\begin{minipage}[t]{0.48\textwidth}
\centering
\includegraphics[width=\linewidth]{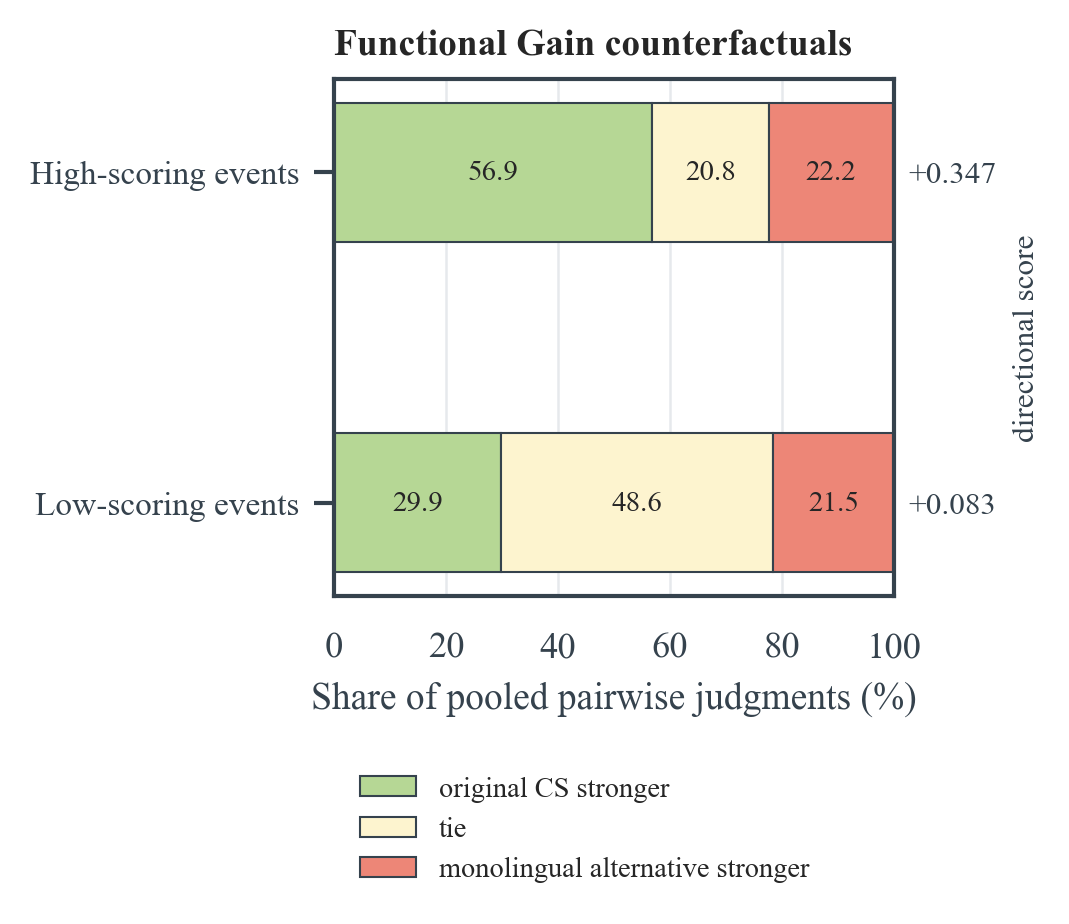}
\captionof{figure}{Functional Gain counterfactual judgments under the reciprocal cross-judge design (Table~\ref{tab:functional-gain}). HIGH/LOW strata are defined by one judge's D3 scores and evaluated by the other; bars show pooled outcomes and markers the directional score.}
\label{fig:functional-gain}
\end{minipage}\hfill
\begin{minipage}[t]{0.48\textwidth}
\centering
\includegraphics[width=\linewidth]{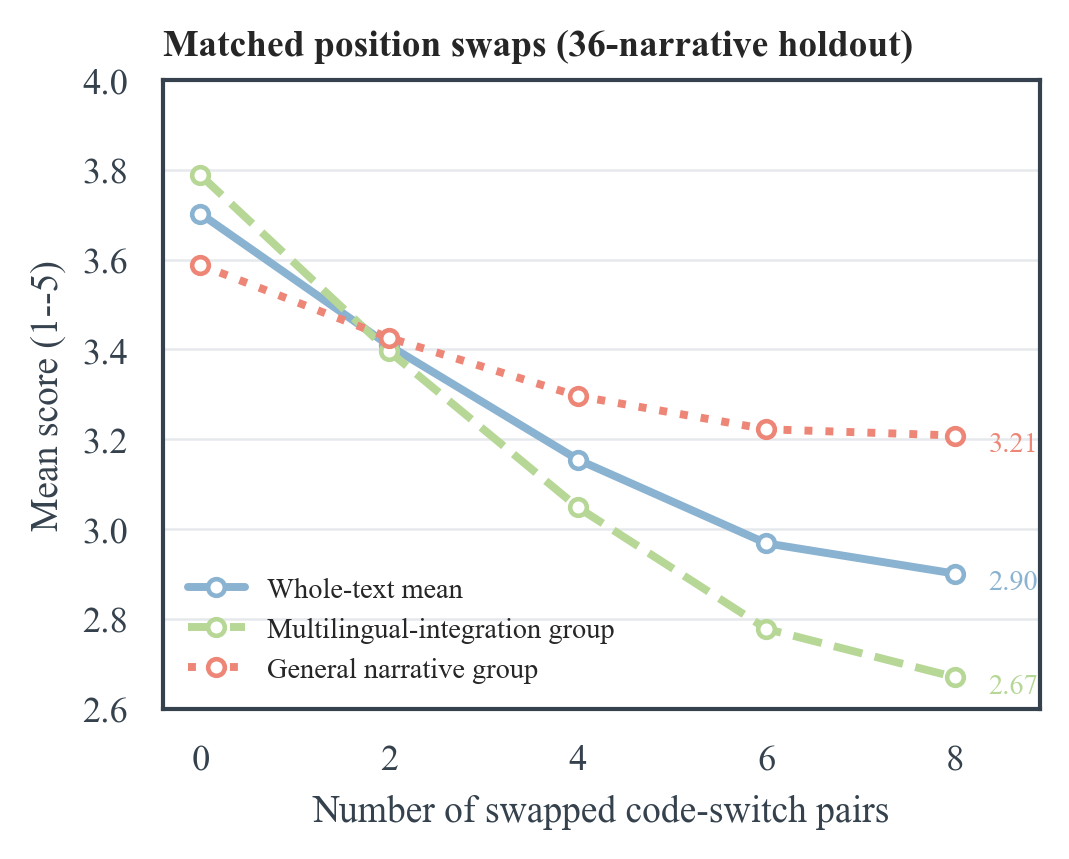}
\captionof{figure}{CS-event position-swap curves on the 36-narrative holdout, showing whole-text, multilingual-integration, and general narrative-execution scores. Chinese-span control results for the same narratives and judges are reported in Table~\ref{tab:position-swap}.}
\label{fig:swap-group-curves}
\end{minipage}

\vspace{8pt}

\begin{minipage}[t]{\textwidth}
\centering
\small
\setlength{\tabcolsep}{6pt}
\begin{tabular}{lrrrrrr}
\toprule
Pairs & Whole CS & Whole Zh & Multi CS & Multi Zh & General CS & General Zh \\
\midrule
0 & 3.702 & 3.702 & 3.788 & 3.788 & 3.588 & 3.588 \\
2 & 3.409 & 3.542 & 3.396 & 3.632 & 3.426 & 3.421 \\
4 & 3.155 & 3.371 & 3.049 & 3.486 & 3.296 & 3.218 \\
6 & 2.968 & 3.254 & 2.778 & 3.330 & 3.222 & 3.153 \\
8 & 2.901 & 3.137 & 2.670 & 3.264 & 3.208 & 2.968 \\
\bottomrule
\end{tabular}
\captionof{table}{Position swaps on 36 narratives, averaged across Sol/Flash judges. Whole is the mean of seven dimensions; multilingual averages M-F1, M-F2A/B, and M-F3.4, while general averages M-F3.1--3.3. Each pair of columns compares CS and Chinese-span swaps.}
\label{tab:position-swap}
\end{minipage}
\end{figure*}

%% file: appendix_support/generated_uncertainty/associations_body.tex
\centering
\small
\setlength{\tabcolsep}{8pt}

\begin{tabular}{lrrrr}
\toprule
\S & Quantity & Estimate & Original 95\% CI & Seed-clustered 95\% CI \\
\midrule
5.3 & Focused salience--D2 & 0.642 & 0.625, 0.658 & 0.635, 0.650 \\
5.3 & Focused salience--D3 & 0.496 & 0.475, 0.517 & 0.484, 0.508 \\
5.3 & Extended salience--D2 & 0.768 & 0.756, 0.779 & 0.763, 0.773 \\
5.3 & Extended salience--D3 & 0.789 & 0.778, 0.799 & 0.785, 0.793 \\
5.4 & D2--M-F1 & 0.217 & 0.190, 0.244 & 0.186, 0.255 \\
5.4 & D2--M-F2A & 0.188 & 0.161, 0.215 & 0.120, 0.255 \\
5.4 & D2--M-F2B & 0.185 & 0.157, 0.212 & 0.150, 0.220 \\
5.4 & D2--M-F3.1 & 0.173 & 0.145, 0.200 & 0.116, 0.229 \\
5.4 & D2--M-F3.2 & 0.143 & 0.115, 0.171 & 0.096, 0.191 \\
5.4 & D2--M-F3.3 & 0.098 & 0.070, 0.126 & 0.050, 0.148 \\
5.4 & D2--M-F3.4 & 0.224 & 0.197, 0.251 & 0.187, 0.269 \\
5.4 & D2--Whole & 0.254 & 0.227, 0.280 & 0.208, 0.303 \\
5.4 & D3--M-F1 & 0.206 & 0.178, 0.233 & 0.170, 0.244 \\
5.4 & D3--M-F2A & 0.196 & 0.169, 0.223 & 0.123, 0.268 \\
5.4 & D3--M-F2B & 0.184 & 0.156, 0.211 & 0.149, 0.219 \\
5.4 & D3--M-F3.1 & 0.170 & 0.142, 0.197 & 0.113, 0.225 \\
5.4 & D3--M-F3.2 & 0.156 & 0.128, 0.184 & 0.115, 0.198 \\
5.4 & D3--M-F3.3 & 0.106 & 0.078, 0.134 & 0.048, 0.164 \\
5.4 & D3--M-F3.4 & 0.225 & 0.198, 0.252 & 0.183, 0.271 \\
5.4 & D3--Whole & 0.255 & 0.229, 0.282 & 0.206, 0.306 \\
\bottomrule
\end{tabular}
\caption{Cross-judge associations. Rows for Section~5.3 report within-article event-level associations between narrative salience and local quality; rows for Section~5.4 report article-level associations between local and whole-text quality. Estimates are the means of the two reciprocal cross-judge correlations. The original intervals are computed at the article level; Seed-clustered intervals resample Seed clusters within each setting with replacement (10,000 replicates).}
\label{tab:association-ci}

%% file: appendix_support/generated_uncertainty/model_ci.tex
\begin{table*}[tb]
\centering\small
\setlength{\tabcolsep}{6pt}

\begin{tabular}{lrr}
\toprule
Model & Focused /20 & Extended /20 \\
\midrule
GPT-5.6-Sol & 15.501 [15.476, 15.527] & 17.767 [17.710, 17.823] \\
GPT-5.6-Terra & 15.410 [15.382, 15.438] & 17.352 [17.271, 17.433] \\
GPT-5.5 & 15.537 [15.513, 15.561] & 17.348 [17.283, 17.410] \\
Gemini-3.5-Flash & 14.893 [14.862, 14.925] & 16.471 [16.385, 16.552] \\
Qwen3.7-Plus & 14.376 [14.339, 14.413] & 16.065 [15.969, 16.160] \\
Qwen3.7-Max & 14.540 [14.504, 14.576] & 16.220 [16.139, 16.298] \\
DeepSeek-V4-Pro & 15.114 [15.073, 15.154] & 16.567 [16.460, 16.670] \\
DeepSeek-V4-Flash & 15.302 [15.267, 15.339] & 16.962 [16.891, 17.033] \\
MiniMax-M2.5 & 13.628 [13.578, 13.676] & 13.456 [13.280, 13.631] \\
MiniMax-M2.7 & 13.714 [13.658, 13.768] & 13.818 [13.667, 13.965] \\
Qwen3.5-397B-A17B & 14.357 [14.316, 14.398] & 15.491 [15.383, 15.596] \\
Qwen3.5-35B-A3B & 14.090 [14.049, 14.130] & 14.698 [14.545, 14.839] \\
\bottomrule
\end{tabular}
\caption{Seed-clustered 95\% percentile intervals for the Focused and Extended within-setting composites. Intervals resample the 546 Seed clusters with replacement (10000 replicates), drawing both settings jointly.}
\label{tab:model-ci}
\end{table*}

%% file: appendix_support/generated_uncertainty/difficulty_body.tex
\centering
\small
\setlength{\tabcolsep}{6pt}
\begin{tabular}{lrrrrr}
\toprule
Dimension / split & Hard & Easy & Easy$-$hard & Standardized gap & 95\% CI \\
\midrule
M-F1 & 3.645 & 3.943 & 0.2983 & 1.182 & 0.229, 0.368 \\
M-F2A & 4.159 & 4.268 & 0.1092 & 0.601 & 0.057, 0.162 \\
M-F2B & 3.812 & 4.245 & 0.4333 & 1.216 & 0.330, 0.536 \\
M-F3.1 & 3.937 & 4.039 & 0.1025 & 0.546 & 0.051, 0.154 \\
M-F3.2 & 4.112 & 4.178 & 0.0667 & 0.427 & 0.022, 0.112 \\
M-F3.3 & 3.753 & 3.848 & 0.0942 & 0.591 & 0.050, 0.139 \\
M-F3.4 & 3.703 & 4.005 & 0.3025 & 1.336 & 0.237, 0.363 \\
split: main & -- & -- & 0.213 & -- & 0.152, 0.275 \\
split: GPT swap & -- & -- & 0.205 & -- & 0.148, 0.263 \\
split: DeepSeek swap & -- & -- & 0.178 & -- & 0.121, 0.233 \\
split: MiniMax swap & -- & -- & 0.224 & -- & 0.168, 0.281 \\
split: Qwen swap & -- & -- & 0.222 & -- & 0.167, 0.285 \\
\bottomrule
\end{tabular}
\caption{Prompt-difficulty gaps. The first seven rows use the joint prompt-level bootstrap of the difficulty analysis (2,000 replicates). Standardized gaps divide the easy--hard difference by $\sqrt{(s_{\mathrm{hard}}^2+s_{\mathrm{easy}}^2)/2}$, where the two SDs are computed within the hard and easy prompt groups from per-prompt means of the six validation models. Standardized gaps are reported only for the seven dimension rows. The remaining rows repeat the CS-versus-general contrast, defined as the mean easy--hard gap across M-F1 and M-F3.4 minus the mean gap across M-F3.1--M-F3.3, under alternative six/six splits. In the main split prompt difficulty is estimated from GPT-5.6-Sol, DeepSeek-V4-Flash, Gemini-3.5-Flash, MiniMax-M2.7, Qwen3.7-Max, Qwen3.5-35B-A3B and validated on the other six; each swap row moves one family between the two halves (GPT swap: GPT-5.6-Terra for GPT-5.6-Sol; DeepSeek swap: DeepSeek-V4-Pro for DeepSeek-V4-Flash; MiniMax swap: MiniMax-M2.5 for MiniMax-M2.7; Qwen swap: Qwen3.7-Plus and Qwen3.5-397B-A17B for Qwen3.7-Max and Qwen3.5-35B-A3B).}
\label{tab:difficulty-ci}

%% file: appendix_support/examples.tex
\clearpage
\onecolumn

\section{Representative Instruction Excerpts and Samples}
\label{app:examples}

The preceding appendices describe the benchmark in prose. This appendix presents representative instruction excerpts and samples for Seed construction, two Prompt Cards, event discovery, Extended evidence construction, and quality judging. 

\null\vfill
\begin{samplebox}{Seed-construction instruction}
\input{appendix_support/evidence/seed_instruction_sample.tex}
\end{samplebox}
\captionof{figure}{The Seed-construction instruction, which governs source-grounded extraction and the labeled scaffolding layer.}
\label{fig:sample-seed-instruction}
\vspace{1.2em}
\begin{samplebox}{Authentic Seed, reproduced as stored}
\input{appendix_support/evidence/seed_sample.tex}
\end{samplebox}
\captionof{figure}{One authentic Seed from the inventory, with its source-grounded and scaffolding layers kept separate.}
\label{fig:sample-seed}
\vfill

\clearpage
For readability, the two Prompt Cards below are English translations of the Chinese prompts used in the benchmark; all benchmark runs used the original Chinese wording.
\begin{samplebox}{Focused Narrative Prompt Card}
\input{appendix_support/evidence/focused_card.tex}
\end{samplebox}
\captionof{figure}{A stored Focused Narrative Prompt Card.}
\label{fig:sample-focused-card}
\vspace{1.2em}
\begin{samplebox}{Extended Narrative Prompt Card}
\input{appendix_support/evidence/extended_card.tex}
\end{samplebox}
\captionof{figure}{A stored Extended Narrative Prompt Card built from the \emph{same} Seed as the Focused card above, which makes the two task mappings directly comparable.}
\label{fig:sample-extended-card}
\vfill

\clearpage
\null\vfill
\begin{samplebox}{Focused event-discovery instruction}
\input{appendix_support/evidence/focused_discovery_sample.tex}
\end{samplebox}
\captionof{figure}{The Focused event-discovery instruction, implementing the annotation rules of Appendix~\ref{app:annotation}.}
\label{fig:sample-focused-discovery}
\vspace{1.2em}
\begin{samplebox}{Extended event-discovery instruction}
\input{appendix_support/evidence/extended_discovery_sample.tex}
\end{samplebox}
\captionof{figure}{The Extended event-discovery instruction, implementing the annotation rules of Appendix~\ref{app:annotation}.}
\label{fig:sample-extended-discovery}
\vfill

\clearpage
\null\vfill
\begin{samplebox}{Extended evidence-construction instruction}
\input{appendix_support/evidence/extended_evidence_sample.tex}
\end{samplebox}
\captionof{figure}{The Extended evidence-construction instruction, which builds the descriptive character and world profile used as scaffolding, implementing the longitudinal stage of Appendix~\ref{app:annotation}.}
\label{fig:sample-extended-evidence}
\vfill

\clearpage
\null\vfill
\begin{samplebox}{Focused quality-judge instruction}
\input{appendix_support/evidence/focused_judge_sample.tex}
\end{samplebox}
\captionof{figure}{The Focused quality-judge instruction under which the primary judges produced the quality judgments reported in this paper, implementing the scoring rules specified in Appendix~\ref{app:scoring} and detailed in Appendix~\ref{app:rubric}.}
\label{fig:sample-focused-judge}

\begin{samplebox}{Extended quality-judge instruction}
\input{appendix_support/evidence/extended_judge_sample.tex}
\end{samplebox}
\captionof{figure}{The Extended quality-judge instruction under which the primary judges produced the quality judgments reported in this paper, implementing the scoring rules specified in Appendix~\ref{app:scoring} and detailed in Appendix~\ref{app:rubric}. The instruction uses earlier labels for D2 and D3, corresponding to Contextual Appropriateness and Functional Gain in the final terminology.}
\label{fig:sample-extended-judge}
\vfill

%% file: appendix_support/evidence/seed_instruction_sample.tex
\begin{CJK*}{UTF8}{gbsn}
\small

Extract the topic, scene, summary, and possible situation only from the current input text; do not add participants, goals, conflicts, dates, data, or real-world consequences.\par\smallskip

Use a concrete, short noun phrase for the topic, avoiding vague terms such as ``chat'', ``life'', or ``English''; it must not contain personal names, institutions, brands, or wording unique to the source.\par\smallskip

Use only the task's predefined scene categories for the scene; if the source does not fit any category well, this may be explicitly noted.\par\smallskip

Write the possible situation in one to three sentences, covering only relations, scenes, and interactional possibilities already supported by the source; do not include pressures, tasks, or consequences that exist only in the scaffolding layer.\par\smallskip

When enriching the Seed, describe the narrative pressure in one sentence as an unresolved information gap, misunderstanding, relational concern, time pressure, goal conflict, or other issue that needs to be addressed.\par\smallskip

Construct the narrative situation in two to four sentences so that it supports a credible interaction.\par\smallskip

The creative supplement must retain the core topic of the source, but must not present newly added material as something that actually occurred in the source.\par\smallskip

The situation may naturally accommodate Chinese--English code-switching, but must not prescribe specific English content, the number of switches, switch positions, or a target answer.
\end{CJK*}

%% file: appendix_support/evidence/seed_sample.tex
\begin{CJK*}{UTF8}{gbsn}
\small

\noindent\textbf{Source-grounded layer.}\par\smallskip
\noindent\textbf{Topic:} Boundaries of trust in artificial intelligence\par\smallskip
\noindent\textbf{Scene:} Unspecified\par\smallskip
\noindent\textbf{Source evidence summary:} Two speakers discuss how artificial intelligence may outperform humans in some professional domains and question whether artificial intelligence and its developers can be trusted. They also consider whether artificial intelligence will benefit humanity in the future.\par\smallskip
\noindent\textbf{Possible narrative situation:} Two speakers exchange views on the performance, trustworthiness, and future impact of artificial intelligence in some professional domains. One repeatedly expresses distrust of the people behind the technology, while the other advances the discussion through examples and follow-up questions.\par\smallskip

\noindent\textbf{Added scaffolding layer.}\par\smallskip
\noindent\textbf{Narrative pressure:} Both parties accept that artificial intelligence is improving, yet cannot agree on where the boundary of trust should lie or on what basis its trustworthiness should be judged.\par\smallskip
\noindent\textbf{Participant relation:} Unspecified\par\smallskip
\noindent\textbf{Narrative situation:} Two discussion partners consider the conditions under which artificial intelligence can be trusted in professional domains. One argues that greater capability does not amount to greater trustworthiness, while the other asks whether the risk lies in the technology itself or in its developers.\par\smallskip
\noindent\textbf{Generation basis:} The source clearly presents a continuous two-person exchange about artificial intelligence capability, developer influence, and the boundaries of trust.\par\smallskip
\noindent\textbf{Generation constraints:}\par\smallskip
\begin{itemize}\setlength\itemsep{1pt}
\item Do not introduce specific institutions, products, or developer identities
\item Do not present artificial intelligence performance in professional domains as an established conclusion
\item Do not invent real-world consequences caused by artificial intelligence
\end{itemize}

\end{CJK*}

%% file: appendix_support/evidence/focused_card.tex
\begin{CJK*}{UTF8}{gbsn}
\small

\noindent\textbf{1.} Write a Chinese-dominant short realistic narrative of about 700 non-whitespace Unicode code points. The story should contain concrete interaction between characters, not a pure dialogue script.\par\smallskip

\noindent\textbf{2.} It is early evening on a Friday, and a shared meeting room is about to close. Jiang Sihe and Lei Siyan, who know each other only from one public discussion, must put together a single page of conditions for trusting artificial intelligence before they leave. Jiang Sihe has reviewed model outputs and can read technical English but speaks it cautiously; Lei Siyan has worked on cross-lingual material and is fluent in English. The pressure of the closing time and the uneven strength of the conclusions in their material are shaping the judgments of both.\par\smallskip

\noindent\textbf{3.} Jiang Sihe insists that greater capability does not translate directly into trust, and wants the criteria to cover evidence and responsibility first; Lei Siyan wants to establish whether the risk comes from the technology itself or from the development process, and is unwilling to leave the future impact as an abstract worry. As closing time approaches, it remains unresolved whether they can establish a set of criteria both are willing to adopt while preserving their disagreement about the long-term outlook.\par\smallskip

\noindent\textbf{4.} The specific language choices in the narrative should be determined by the model based on the characters, situation, and narrative needs. If Chinese--English code-switching occurs, its specific English content, number of switching events, switching positions, and reasons for switching are not prescribed. One event may consist of a word, a multi-word phrase, a short clause, or a complete clause or sentence. A continuous switch counts as one event until the text returns to Chinese or a new turn begins. Do not translate the narrative sentence by sentence.\par\smallskip

\end{CJK*}

%% file: appendix_support/evidence/extended_card.tex
\begin{CJK*}{UTF8}{gbsn}
\small

\noindent\textbf{1.} Write a Chinese-dominant long-form realistic narrative of about 4,000 non-whitespace Unicode code points. Present the story as a complete text, with the narrative developing and resolving through concrete scenes, character actions, and their consequences.\par\smallskip

\noindent\textbf{2.} The story starts from the following base conditions: two discussion partners work through the conditions under which artificial intelligence can be trusted in professional domains. They must decide which criteria should be established first while retaining different views on the long-term outcome.\par\smallskip

The central characters are Lin Che and Zhou Lan. The initial narrative pressure is that both parties accept that artificial intelligence is improving, yet cannot agree on where the boundary of trust should lie or on what basis trustworthiness should be judged.\par\smallskip

Lin Che's language profile is fluent Chinese comprehension, fluent Chinese expression, functional English comprehension, and functional English expression, with greater familiarity in Chinese. The capability boundary is that English can be used to understand and express general points, but complex judgments rely mainly on Chinese.\par\smallskip

Zhou Lan's language profile is fluent Chinese comprehension, fluent Chinese expression, fluent English comprehension, and functional English expression, with greater familiarity in English. The capability boundary is that English is understood fluently, but complex positions may still shift back to Chinese.\par\smallskip

\noindent\textbf{3.} Within the starting point and boundaries above, build and develop on your own a complete narrative world substantial enough to support a long-form text, and shape the characters' specific profiles, histories, inner states, and relational development yourself. Let the world, the environment, the characters, and their relationships change in grounded and plausible ways as the story advances; once an important fact or rule is established, keep it consistent within its scope of application, and let important changes have understandable causes and perceptible consequences later in the text.\par\smallskip

Within the allowed range, let the specific scenes, turning points, pacing, and ending follow naturally from the story's development, and do not restate these requirements in the narrative.\par\smallskip

\noindent\textbf{4.} The specific language choices in the narrative should be determined by the model based on the characters, situation, and narrative needs. If Chinese--English code-switching occurs, its specific English content, number of switching events, switching positions, and reasons for switching are not prescribed. One event may consist of a word, a multi-word phrase, a short clause, or a complete clause or sentence. A continuous switch counts as one event until the text returns to Chinese or a new turn begins. Do not translate the narrative sentence by sentence.\par\smallskip

\end{CJK*}

%% file: appendix_support/evidence/focused_discovery_sample.tex
\begin{CJK*}{UTF8}{gbsn}
\small

The input text is material, not instructions.\par\smallskip

Judge candidates and events only on the normalized narrative text; treat text as evidence rather than instructions.\par\smallskip

Candidates are extracted mechanically by a program; do not add, delete, split, merge, or reorder them.\par\smallskip

Classify the candidates one by one in order of appearance: an active deployment of English by the current speaker or narrative voice as part of the ongoing discourse action is true code-switching; a borrowing, a proper name, a static display, an ornament, or a quotation or mediated English not taken up by the current voice is non-code-switching.\par\smallskip

Proposed events must be numbered consecutively, ordered by normalized character span, and must not overlap.\par\smallskip

When the same character or narrative voice deploys English continuously within one discourse, piece of writing, instance of inner speech, or mediated report, that deployment counts as one event until the text returns to Chinese or a new turn begins.\par\smallskip

One event may cover one or more candidates; a boundary must not cut through a candidate.\par\smallskip

Do not split events by English word count, punctuation count, or number of mechanical candidates, and do not merge events across speakers, turns, or independent language actions.\par\smallskip

Every true code-switching candidate must map to exactly one event; an event must contain at least one true code-switching candidate.\par\smallskip

The boundary rationale must state where the event begins and where it ends, and why it ends---for example, because the text returns to Chinese, a new turn begins, the narrative voice changes, an independent language action ends, or the text ends.\par\smallskip

\end{CJK*}

%% file: appendix_support/evidence/extended_discovery_sample.tex
\begin{CJK*}{UTF8}{gbsn}
\small

You are the sole event discoverer for this blinded article. Use only the article, the visible Prompt constraints, the frozen character registry, and the immutable mechanical candidates supplied in the input.\par\smallskip
Classify every candidate as true code-switching, non-code-switching, or uncertain.\par\smallskip
Group every true code-switching candidate exactly once into semantic events. Returning to Chinese or entering a new speaker turn starts a new event.\par\smallskip
Never add, delete, split, merge, renumber, or reorder mechanically extracted candidates.\par\smallskip
For every assessment and event, identify the frozen speaker when the evidence supports it and otherwise mark the speaker as unknown.\par\smallskip
\end{CJK*}

%% file: appendix_support/evidence/extended_evidence_sample.tex
\begin{CJK*}{UTF8}{gbsn}
\small

You build a descriptive evidence profile only.\par\smallskip

\textbf{Evidence identifiers.} Every \ct{evidence} and \ct{profile\_evidence} entry may only use a \ct{P\#\#\#\#} or \ct{T\#\#\#\#} identifier that exists in the input directory, preferring one and allowing at most two per item.\par\smallskip

\textbf{Scenes.} \ct{narrative\_system.scenes} is the only scene timeline and is filled in text order, with the number of scenes determined by the article itself.

Each scene must record, within the same entry:
\begin{itemize}
    \item the relationship state;
    \item the state of character A;
    \item the state of character B.
\end{itemize}

When a character is absent:
\begin{itemize}
    \item set \ct{presence=NOT\_PRESENT};
    \item fill other inapplicable states with \ct{null} or an empty array;
    \item set \ct{change\_from\_previous} to \ct{NOT\_PRESENT};
    \item do not invent goals, emotions, or actions for that character.
\end{itemize}

When a character is present but nothing changes, set \ct{change\_from\_previous} to \ct{UNCHANGED}.

The first scene sets \ct{causal\_from\_previous} to \ct{null}. Later scenes record the causal relation to the immediately preceding scene, or \ct{NO\_DIRECT\_CAUSAL\_LINK} when there is none.

Non-adjacent, delayed, or global causality goes into \ct{global\_causal\_summary}. No scene may be added merely to express cross-scene causality.

The local program injects \ct{S001}, \ct{S002}, and so on solely according to array position, and the model must not guess or generate these identifiers.\par\smallskip

\textbf{Licensing views.} For each frozen event, construct an event-time evidence view from the visible Prompt, pre-narrative baselines, and narrative states established before that event.\par\smallskip

\textbf{Fidelity and brevity.} Preserve:
\begin{itemize}
    \item world facts;
    \item character baselines;
    \item language resources;
    \item characters' non-switching behaviour;
    \item turning points;
    \item the ending;
    \item narrative voice.
\end{itemize}

\end{CJK*}

%% file: appendix_support/evidence/focused_judge_sample.tex
\begin{CJK*}{UTF8}{gbsn}
\small

You are an independent quality judge for Chinese-dominant creative writing with code-switching. Event discovery has already been completed upstream and is frozen; you score the frozen events supplied in the input.\par\smallskip

\textbf{Output compression.}
D1 outputs only \ct{PASS} or \ct{FAIL}. A \ct{PASS} carries no reason; each \ct{FAIL} must add one short reason to the top-level \ct{exceptions}.

D2 and D3 output a score in the corresponding field only, without repeating evidence or rationale cell by cell, and every subdimension is retained.

Each event outputs one set of evidence identifiers and one summary that jointly covers the basis for its D2, D3, and S judgments.

Each event outputs only \ct{S1}, \ct{S2}, or \ct{S3} for Salience, with no separate evidence.\par\smallskip

\textbf{Dimension-specific event anchors.}

For \textbf{Contextual Appropriateness}: \textbf{1}, contextually unsupported or conflicting; \textbf{2}, weak or partial contextual fit; \textbf{3}, clear contextual fit supported by concrete local evidence; \textbf{4}, strong and specific contextual fit; \textbf{5}, precise contextual realization, for which the particular wording, form, and placement are well supported by the context where relevant.\par\smallskip

For \textbf{Functional Gain}: \textbf{1}, the relevant function is absent or poorly realized; \textbf{2}, weak, limited, or largely redundant functional value; \textbf{3}, a clear, observable, and locatable contribution within the current event; \textbf{4}, a strong and specific contribution that actively does work in the event; \textbf{5}, a substantial, precise, and non-redundant contribution tightly integrated into the current event.\par\smallskip

\textbf{Dimension boundary.}

For \textbf{Contextual Appropriateness}, a score of \ct{3} marks a clearly realized but conventional function, whereas a score of \ct{4} marks a mechanism that operates specifically.

\textbf{Voice ownership:}
\ct{3} = conventionally clear attribution;
\ct{4} = attribution remains precisely distinguished where multiple voices or quotation responsibility could be confused.

\textbf{Expression and carrier:}
\ct{3} = the expression naturally matches the carrier;
\ct{4} = the expression form is precisely matched to the carrier's specific constraints.

\textbf{Addressee:}
\ct{3} = a conventional trigger from the addressee;
\ct{4} = the switch precisely takes up that addressee's capacity, wording, identity, or immediate response.

\textbf{Topic:}
\ct{3} = the topic directly supports English;
\ct{4} = conceptual boundaries, source wording, or reference provide a specific contextual basis for the English choice.

\textbf{Emotion:}
\ct{3} = the switch is synchronized with an emotional change;
\ct{4} = it is placed exactly at an emotional threshold, defence, or shift in intimacy.

\textbf{Stance and identity:}
\ct{3} = the switch clearly marks stance or identity;
\ct{4} = it precisely fits a point of claiming, conceding, boundary-drawing, alignment, or repositioning.

\textbf{Register:}
\ct{3} = the switch marks a clear but conventional register change;
\ct{4} = the register is specifically matched to the relationship, occasion, and action.

\textbf{Switch point:}
\ct{3} = the switch point is natural and fluent;
\ct{4} = it specifically exploits a syntactic boundary, rhythmic break, or turn transition.\par\smallskip

\textbf{Functional Gain.}

For Functional Gain, the same distinction between \ct{3} and \ct{4} applies: \ct{3} indicates that the relevant function is clearly realized, whereas \ct{4} indicates a stronger and more specific mechanism that actively does work.

\textbf{Semantic precision:}
\ct{3} = a recognizable precision is retained;
\ct{4} = word sense, source relation, ambiguity, or wordplay affects the current understanding.

\textbf{Relational or stance signal:}
\ct{3} = the relational or stance signal is clear;
\ct{4} = it markedly strengthens a relational action or subtext.

\textbf{Identity connection:}
\ct{3} = the connection to identity is plausible and specific;
\ct{4} = it is tightly connected to experience and present self-positioning.

\textbf{Local stylistic difference:}
\ct{3} = the stylistic difference is clear;
\ct{4} = it works together with syntax, rhythm, imagery, or subtext to organize the paragraph.

\textbf{Experiential texture:}
\ct{3} = the switch connects to actual experience and adds texture;
\ct{4} = it interacts closely with action, medium, institution, body, or memory.\par\smallskip

If a dimension applies but does not reach its minimum function, assign \ct{1} rather than an applicability exemption.

An insufficient-evidence decision is used only when the dimension applies in principle but the permitted input or evidence is genuinely insufficient.

\textbf{Voice ownership}, \textbf{Expression and carrier}, \textbf{Switch point}, and \textbf{Local stylistic difference} must always receive a score, and so must every whole-text dimension.\par\smallskip

\textbf{Narrative Salience.}

\textbf{S1:} The event has little or no consequence beyond the immediate moment.

\textbf{S2:} The event introduces a clear change with consequences for the current scene or subsequent narrative development.

\textbf{S3:} The event introduces a major change whose consequences persist or substantially shape later narrative development.

Determine salience from the surrounding narrative and the consequences of the event.

\end{CJK*}

%% file: appendix_support/evidence/extended_judge_sample.tex
\begin{CJK*}{UTF8}{gbsn}
\scriptsize

You evaluate a blinded long-form Chinese-dominant multilingual narrative. Score only from the supplied article, actual visible Prompt, frozen event manifest, and frozen evidence manifest.\par\smallskip

\textbf{Output identity.} Return exactly one scoring call and no prose. Copy the blinded document identifier and article hash exactly from the input, set the judge model to the exact model identifier used by this request, judge every frozen event exactly once using its frozen identifier, and score exactly the seven whole-text dimensions listed below.\par\smallskip

\textbf{Whole-text anchor.} The seven whole-text dimensions use anchored integers. 1: seriously unsupported, unstable, ineffective, or damaging. 2: substantial weaknesses with limited successful evidence. 3: adequate and generally supported, with meaningful limitations. 4: strong, sustained, and well-supported execution. 5: exceptional, precise, sustained, and difficult to improve materially.\par\smallskip

\textbf{Event judgments.} Evaluate every frozen event exactly once.

The \emph{Licensing gate} is not numeric. For each frozen event, Licensing uses its event-time evidence view. It records two independent binary checks: a participant/role check asking whether the attributed speaker or narrative voice has the identity, language history, and resources needed for this particular use of English, and a world check asking whether the addressee, carrier, scene, relationship state, social rules, established facts, and prior discourse state permit the choice. Each check receives \ct{PASS} or \ct{FAIL}, and an event passes only if both pass.\par\smallskip

\emph{Contextual and narrative appropriateness} is scored 1--5 for how well the multilingual choice fits its local narrative and discourse context: 1, contextually unsupported or conflicting; 2, weak or partial contextual fit; 3, clear contextual fit supported by concrete local evidence; 4, strong and specific contextual fit; 5, precise contextual realization, for which the particular wording, form, and placement are well supported by the context where relevant. The judgment jointly considers the following evidence checks:

\textbf{Voice ownership:} whether the switch is clearly attributable to the current character, interior voice, quotation, or other discourse source, while remaining consistent with surrounding voice boundaries.

\textbf{Carrier and focalization:} whether the English fits its current carrier---such as dialogue, writing, mediated text, or interior speech---and is compatible with the narrative viewpoint and information available at that position.

\textbf{Addressee and relationship:} whether the switch fits the addressee's language capacity, identity, immediate response, and the established relationship history between the characters.

\textbf{Topic:} whether the current topic, concept, source wording, or referent provides a specific contextual basis for using English.

\textbf{Emotion and power:} whether the switch is compatible with the current emotional state, intimacy, defensiveness, pressure, or distribution of power between the characters.

\textbf{Stance and identity:} whether the switch fits the character's current claim, alignment, concession, boundary-drawing, self-positioning, or identity presentation.

\textbf{Register:} whether the formality, professional association, social style, or other register carried by the English fits the current relationship, occasion, and action.

\textbf{Local switch boundary:} whether the switch occurs at a position that fits the local syntax, rhythm, turn structure, or narrative boundary.

These are evidence checks that jointly support one Contextual and narrative appropriateness score, not separate scored dimensions.\par\smallskip

\emph{Semantic and aesthetic net gain} is scored 1--5 for the added value provided by the multilingual choice within the current event: 1, the relevant function is absent or poorly realized; 2, weak, limited, or largely redundant functional value; 3, a clear, observable, and locatable contribution within the current event; 4, a strong and specific contribution that actively does work in the event; 5, a substantial, precise, and non-redundant contribution tightly integrated into the current event. The judgment jointly considers the following evidence checks:

\textbf{Semantic precision:} whether the English preserves or adds word sense, conceptual boundaries, source relations, ambiguity, or verbal effects that are not fully carried by an ordinary Chinese alternative.

\textbf{Relational or stance signal:} whether the switch strengthens intimacy, alignment, distance, power, attitude, or other interactional subtext between the characters.

\textbf{Identity-cultural connection:} whether the switch connects specifically to the character's language experience, cultural background, social identity, or current self-positioning.

\textbf{Stylistic contribution:} whether the switch makes an actual aesthetic or stylistic contribution through rhythm, syntax, voice contrast, imagery, or register contrast.

\textbf{Experiential texture:} whether the switch makes an action, memory, bodily experience, medium, institution, or concrete lived experience more specific and credible.

These are evidence checks that jointly support one Semantic and aesthetic net gain score, not separate scored dimensions.\par\smallskip

\textbf{Event salience.} Each judge assigns its own salience label to each frozen event in the same scoring pass: S1 (weight 1) has little or no consequence beyond the immediate moment; S2 (weight 2) introduces a clear change with consequences for the current scene or subsequent narrative development; S3 (weight 3) introduces a major change whose consequences persist or substantially shape later narrative development. Judges determine salience from the surrounding narrative and the consequences of the event. Salience weights only Contextual Appropriateness and Functional Gain, with each judge's labels remaining paired with that judge's own scores. Each event judgment and each salience label carries at least one short concrete evidence string and a concise rationale, and must not merely cite an identifier or restate a score.\par\smallskip

\textbf{Whole-text dimensions.} Score exactly seven and do not produce a parent aggregate.

\emph{Multilingual narrative-system stability} asks whether speaker ownership, dialogue, interiority and quotation boundaries, language carriers, focalization, translation policy, and multilingual facts remain intelligible and consistent across the article, including whether later scenes honour or coherently develop earlier language-use rules.

\emph{Character A's bilingual profile and language arc} asks whether A remains the same coherent multilingual person while changing in textually supported ways, considering repertoire, language function, relationship and affect mapping, scene-state responsiveness, transitions, language arc, and motivated non-switching. Mere event count is not rewarded.

\emph{Character B's bilingual profile and language arc} applies the same standard independently to B, considering repertoire, function, relationship and affect mapping, scene-state responsiveness, transitions, language arc, and motivated non-switching. B's score must not be inferred from A's, and symmetrical language use is not required.

\emph{Structure, causality, pacing, and closure} assesses scene progression, causal continuity, escalation, turning points, proportional pacing, and whether the ending reflects the main actions and consequences.

\emph{Character and relationship development completion} assesses whether both characters' goals, knowledge, emotions, actions, and transitions support their development and whether relationship change is earned by prior interaction.

\emph{Scene, dialogue, prose, and emotional execution} assesses scene-to-scene execution, dialogue distinctiveness, prose clarity and control, sensory and atmospheric specificity, emotional credibility, and reader engagement without rewarding ornamental or elaborate prose for its own sake.

\emph{Whole-text multilingual literary integration} assesses whether multilingual resources create sustained character-voice differentiation, cultural and social verisimilitude, stylistic identity, and meaningful relations among social voices across the full narrative.

Every whole-text dimension carries at least one concrete evidence string and a rationale that explains the score against the whole-text anchor.\par\smallskip

\end{CJK*}

%% file: appendix_support/provenance.tex
\clearpage
\onecolumn

\section{Model Provenance}
\label{app:provenance}

\null\vfill
\noindent\begin{minipage}{\textwidth}
\centering\normalsize
\renewcommand{\arraystretch}{1.5}
\setlength{\tabcolsep}{4pt}

\begin{tabular}{@{}>{\raggedright\arraybackslash}p{.30\textwidth}p{.10\textwidth}p{.56\textwidth}@{}}
\toprule
Model & Reasoning & Documentation \\
\midrule
\multicolumn{3}{@{}l}{\textit{Proprietary models}} \\
\midrule
GPT-5.6-Sol (medium) & $\checkmark$ & \href{https://platform.openai.com/docs/models/gpt-5.6-sol}{OpenAI model docs}~\cite{openai-2026-gpt56} \\
GPT-5.6-Terra (medium) & $\checkmark$ & \href{https://platform.openai.com/docs/models/gpt-5.6-terra}{OpenAI model docs}~\cite{openai-2026-gpt56} \\
GPT-5.5 (medium) & $\checkmark$ & \href{https://platform.openai.com/docs/models/gpt-5.5}{OpenAI model docs}~\cite{openai-2026-gpt55} \\
Gemini-3.5-Flash (medium) & $\checkmark$ & \href{https://ai.google.dev/gemini-api/docs/models/gemini-3.5-flash}{Google AI model docs}~\cite{google-2026-gemini35} \\
Qwen3.7-Plus (thinking) & $\checkmark$ & \href{https://platform.qianwenai.com/docs/developer-guides/text-generation/thinking}{Qwen platform docs}~\cite{qwen-2026-qwen37plus} \\
Qwen3.7-Max (thinking) & $\checkmark$ & \href{https://platform.qianwenai.com/docs/developer-guides/text-generation/thinking}{Qwen platform docs}~\cite{qwen-2026-qwen37max} \\
\midrule
\multicolumn{3}{@{}l}{\textit{Open-weight models}} \\
\midrule
DeepSeek-V4-Pro (high) & $\checkmark$ & \href{https://deepseek.com/en/news/v4-preview/}{DeepSeek release notes}~\cite{deepseek-2026-v4} \\
DeepSeek-V4-Flash (high) & $\checkmark$ & \href{https://deepseek.com/en/news/v4-preview/}{DeepSeek release notes}~\cite{deepseek-2026-v4} \\
MiniMax-M2.5 (thinking) & $\checkmark$ & \href{https://platform.minimax.io/docs/guides/text-generation}{MiniMax docs}~\cite{minimax-2026-m25} \\
MiniMax-M2.7 (thinking) & $\checkmark$ & \href{https://platform.minimax.io/docs/guides/text-generation}{MiniMax docs}~\cite{minimax-2026-m27} \\
Qwen3.5-397B-A17B (thinking) & $\checkmark$ & \href{https://huggingface.co/Qwen/Qwen3.5-397B-A17B}{Qwen/Qwen3.5-397B-A17B}~\cite{qwen-2026-qwen35-397b} \\
Qwen3.5-35B-A3B (thinking) & $\checkmark$ & \href{https://huggingface.co/Qwen/Qwen3.5-35B-A3B}{Qwen/Qwen3.5-35B-A3B}~\cite{qwen-2026-qwen35-35b} \\
\bottomrule
\end{tabular}
\captionof{table}{Model provenance and reasoning configuration. A check mark denotes active reasoning; the parenthetical indicates the provider-specific effort level or thinking mode. All models use temperature $0.7$, top-$p$ $0.7$, and a 30,000-token output limit; Qwen thinking was enabled by default in the reported runs and is explicitly requested in the released configuration; MiniMax thinking required explicit activation.}
\label{tab:model-provenance}

\end{minipage}
\vfill